\documentclass[10pt,twocolumn,letterpaper,table,xcdraw]{article}

\usepackage{cvpr}              

\usepackage{booktabs}
\usepackage{amssymb}
\usepackage{array}
\usepackage{epsfig}
\usepackage{amsmath}
\usepackage{amssymb}
\usepackage{multicol}
\usepackage{multirow}
\usepackage{nccmath}
\usepackage{pifont}
\usepackage{adjustbox}
\usepackage{enumitem}
\usepackage{caption}
\usepackage{booktabs}
\usepackage{ragged2e}
\usepackage{comment}
\usepackage{color}
\usepackage{float}
\usepackage{tcolorbox}

\usepackage{wrapfig}
\usepackage{algorithm}
\usepackage{algorithmic}
\definecolor{cvprblue}{rgb}{0.21,0.49,0.74}
\usepackage[pagebackref,breaklinks,colorlinks,allcolors=cvprblue]{hyperref}

\definecolor{darkergreen}{RGB}{19,168,33}
\newcommand{\greencheck}{{\color{darkergreen}\ding{51}}}
\newcommand{\redcross}{{\color{red}\ding{55}}}

\newcolumntype{x}[1]{>{\centering\arraybackslash\hspace{0pt}}p{#1}}

\def\paperID{9650} 
\def\confName{CVPR}
\def\confYear{2025}

\newcommand{\method}{\textsc{GenManip}\xspace}
\newcommand{\benchmark}{\textsc{\method-Bench}\xspace}

\title{\method: LLM-driven Simulation for Generalizable\\Instruction-Following Manipulation}

\author{%
    Ning Gao$^{1,2*}$, Yilun Chen$^{1*\ddagger}$, Shuai Yang$^{1,3*}$, Xinyi Chen$^{1,4*}$, Yang Tian$^{1}$, Hao Li$^{1}$,\\ Haifeng Huang$^{1,3}$, Hanqing Wang$^{1}$, Tai Wang$^{1}$, Jiangmiao Pang$^{1\dag}$ \\ 
    \vspace{-3mm} \\
    $^1$Shanghai AI Laboratory,
    $^2$Xi'an Jiaotong University, \\
    $^3$Zhejiang University, 
    $^4$Nanjing University
}

\begin{document}
\maketitle

\renewcommand{\thefootnote}{}
\footnotetext{$^*$ Equal contribution.}
\footnotetext{$^\ddagger$ Project lead.}
\footnotetext{$^\dag$ Corresponding author.}


\begin{figure*}[t]
    \centering
    \includegraphics[width=1.\linewidth]{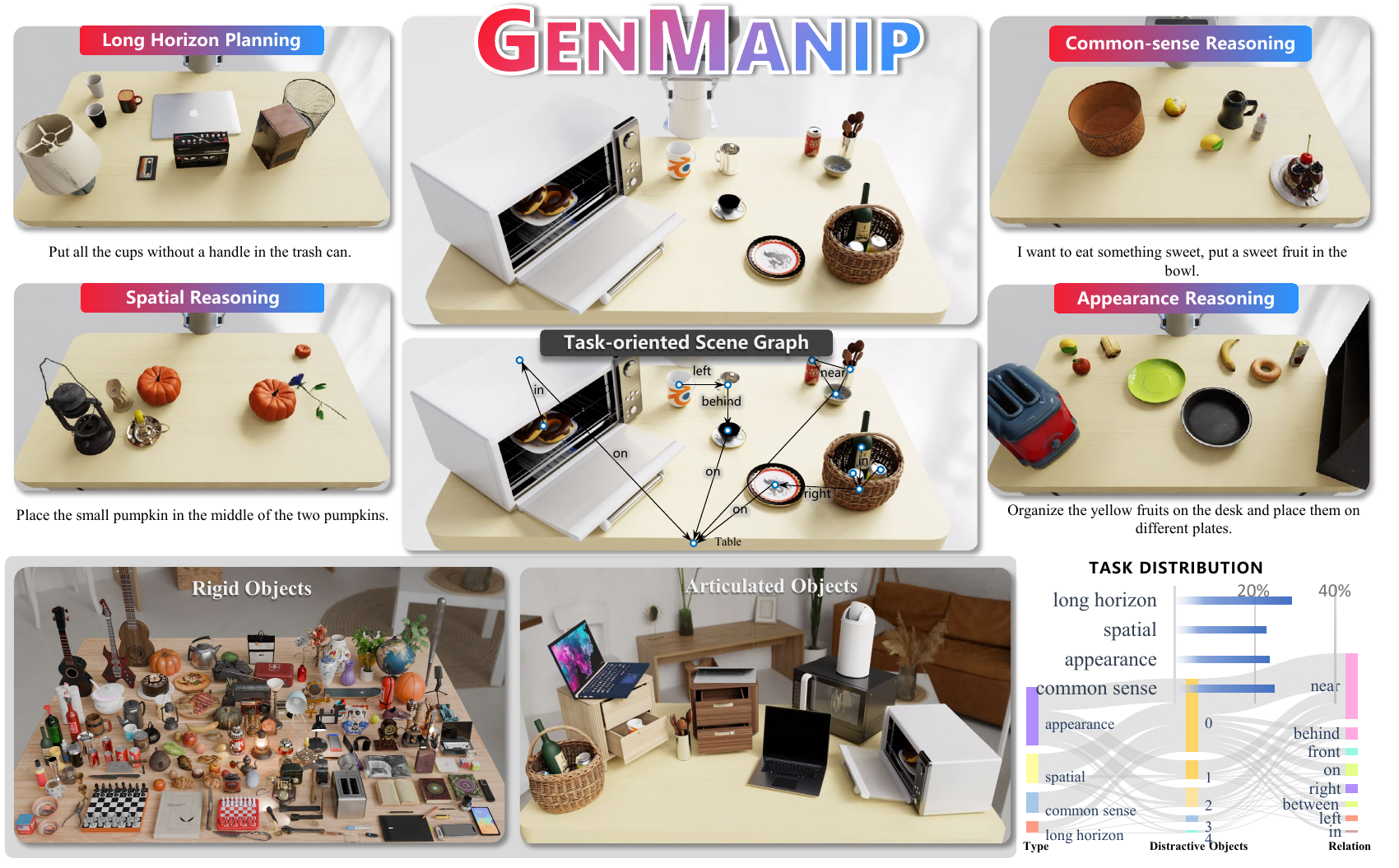}
    \caption{\textbf{\method task scenarios and assets.} \method covers four main task types: long-horizon planning, spatial, commonsense, and appearance-based reasoning. We use the \textit{task-oriented scene graph} (ToSG) to describe each scenario. A subset of rigid (200/10K) and articulated (8/100) objects is included. Task distribution across training and benchmark sets is shown in the bottom-right table.}
    \label{fig:teaser}
\end{figure*}

\begin{abstract} Robotic manipulation in real-world settings remains challenging, especially regarding robust generalization. Existing simulation platforms lack sufficient support for exploring how policies adapt to varied instructions and scenarios. Thus, they lag behind the growing interest in instruction-following foundation models like LLMs, whose adaptability is crucial yet remains underexplored in fair comparisons. To bridge this gap, we introduce \textbf{\method}, a realistic tabletop simulation platform tailored for policy generalization studies. It features an automatic pipeline via LLM-driven \textit{task-oriented scene graph} to synthesize large-scale, diverse tasks using 10K annotated 3D object assets. To systematically assess generalization, we present \textbf{\benchmark}, a benchmark of 200 scenarios refined via human-in-the-loop corrections. We evaluate two policy types: (1) modular manipulation systems integrating foundation models for perception, reasoning, and planning, and (2) end-to-end policies trained through scalable data collection. Results show that while data scaling benefits end-to-end methods, modular systems enhanced with foundation models generalize more effectively across diverse scenarios. We anticipate this platform to facilitate critical insights for advancing policy generalization in realistic conditions. All code will be available at \href{https://genmanip.axi404.top/}{project page}.
\end{abstract}


\section{Introduction}
\label{sec:intro}

Policy generalization is a core challenge within the field of robotic manipulation. Scholars have identified two popular approaches to enhance the generalizability of models. One streams~\cite{moka, copa, google2024pivot, voxposer, vila} involves the hierarchical use of established foundation models, which utilize multi-modal large language models~\cite{GPT-4, gemini, claude, chen2024internvl, sam, groundedsam} (MLLMs), and 2D foundation models~\cite{sam, clip, ovdet, groundedsam} for perception, reasoning, or task and motion planning.
Another strategies emphasizes end-to-end imitation learning, including recent popular vision-language-action models (VLAs) pretrained with large-scale vision-language models (VLMs).
trained on large-scale robotic datasets~\cite{openxembodiment} and accomplish downstream tasks in a zero-shot or downstream fine-tuning manner.
However, benchmarking both policy types for generalization in real-world environments remains challenging due to the high cost and difficulty of replicating exact scenarios consistently. Therefore, high-quality simulation platforms are crucial for systematic policy evaluation.

Previous robotic manipulation benchmarks~\cite{rlbench, jiang2022vima, mees2022calvin, shridhar2020alfred, Libero} often evaluate robots in simplified or controlled environments, emphasizing closed-loop training and evaluation, while sacrificing realism and scenario diversity. Recent efforts have introduced more realistic, everyday, and long-horizon tasks~\cite{behavior1k} and leveraged automatic task generation via LLMs~\cite{gensim, robocasa} or generative models~\cite{liu2023zero}. Nonetheless, these platforms either lack scalable frameworks for generating diverse tasks or comprehensive robotic demonstration data. RoboCasa~\cite{robocasa}, one of the most relevant works, significantly advances realism and task diversity but lacks a standardized evaluation benchmark in diverse instructions and scenarios. The evaluation still primarily employs randomized layouts, limiting systematic evaluation in realistic, everyday scenarios. 

To address these limitations, we introduce \textbf{\method}, a realistic, large-scale tabletop simulation platform tailored for evaluating generalist robots across diverse scenarios and instructions. As illustrated in \cref{fig:teaser}, \method features a rich collection of diverse 3D object assets, a standardized paradigm called the \textit{task-oriented scene graph} (ToSG), and a LLM-driven automatic task generation pipeline. The ToSG serves as a scalable and structured representation of task-related objects, similar in spirit to the natural language adaptation of PDDL~\cite{pddl}, explicitly encoding both object-level and spatial relationships as nodes and edges in the graph. Importantly, the ToSG structure is fully compatible with LLMs, facilitating flexible access to privileged scenario information.

\begin{itemize}[leftmargin=*]
    \item \textbf{Versatile task generation}: 
    ToSG enables diverse task synthesis focusing on (1) understanding intra-object properties (\eg, appearance and shape) and physical properties (\eg, scale, mass), (2) reasoning about inter-object spatial relationships, (3) integrating common-sense knowledge via semantics, and (4) performing long-horizon task execution.
    
    \item \textbf{Controllable layout construction}:
    Unlike purely randomized layouts, ToSG allows explicit control over scenario configurations, substantially reducing ambiguity during task generation, data collection, and policy evaluation.
    
    \item \textbf{Task completion evaluation}: 
    Evaluation allows transforms the final layouts back into ToSG format, enabling straightforward and systematic logical comparisons with target states for accurate task success assessment.
\end{itemize}

\begin{table*}[ht]
\setlength{\tabcolsep}{0pt}
    \centering
    \caption{\textbf{Comparison with existing tabletop manipulation frameworks  used in the robot learning literature.}}
    \resizebox{\textwidth}{!}{%
        \begin{tabular}{ll|cccccccccccc}
        \toprule

        \multicolumn{2}{c|}{\textbf{Simulation Platform}}
        & \begin{tabular}[c]{@{}c@{}}\rotatebox{30}{\textbf{\method}}\end{tabular}
        & \begin{tabular}[c]{@{}c@{}}\rotatebox{30}{\textbf{RoboCasa~\cite{robocasa}}}\end{tabular}
        & \begin{tabular}[c]{@{}c@{}}\rotatebox{30}{\textbf{RLBench~\cite{rlbench}}}\end{tabular} 
        & \begin{tabular}[c]{@{}c@{}}\rotatebox{30}{\textbf{VLMBench~\cite{zheng2022vlmbench}}}\end{tabular} 
        & \begin{tabular}[c]{@{}c@{}}\rotatebox{30}{\textbf{CALVIN~\cite{mees2022calvin}}}\end{tabular} 
        & \begin{tabular}[c]{@{}c@{}}\rotatebox{30}{\textbf{LIBERO~\cite{Libero}}}\end{tabular} 
        & \begin{tabular}[c]{@{}c@{}}\rotatebox{30}{\textbf{Ravens~\cite{zeng2021transporter}}}\end{tabular} 
        & \begin{tabular}[c]{@{}c@{}}\rotatebox{30}{\textbf{Arnold~\cite{gong2023arnold}}}\end{tabular} 
        & \begin{tabular}[c]{@{}c@{}}\rotatebox{30}{\textbf{VIMA~\cite{jiang2022vima}}}\end{tabular} 
        & \begin{tabular}[c]{@{}c@{}}\rotatebox{30}{\textbf{Colosseum~\cite{pumacay2024colosseum}}}\end{tabular} 
        & \begin{tabular}[c]{@{}c@{}}\rotatebox{30}{\textbf{ManiSkill2~\cite{maniskill2}}}\end{tabular} 
        
        \\ \midrule
        \multicolumn{2}{l|}{Engine} & Isaac Sim & Mujuco & RLBench & RLBench & PyBullet & PyBullet & Isaac Sim & Ravens & RLBench & RLBench & SAPIEN \\ \midrule
        \multirow{2}{*}{Assets} & \#Unique Objects & 10K & 2.5K & 28 & 30 & 24 & x & - & 40 & 29 & - & 2.1K \\
        & VL annotations & \greencheck & \redcross & \redcross & \redcross & \redcross & \redcross & \redcross & \redcross & \redcross & \redcross & \redcross \\ 
        \midrule
        \multirow{4}{*}{Benchmark} & Human-in-the-loop & \greencheck & \redcross & \redcross & \redcross & \redcross & \greencheck &  \redcross & \redcross & \redcross & \redcross & \redcross \\
        & \#Tasks & 200 & 100 & 100 & - & 34 & 130 & - & - & 17 & 20 & 20 \\
        & Multi-aspect Evaluation  & \greencheck & \redcross & \redcross & \redcross & \redcross & \greencheck & \redcross & \redcross & \redcross & \greencheck & \greencheck \\ 
        & Multi-step Evaluation & \greencheck & \greencheck & \redcross & \redcross & \redcross & \redcross & \redcross & \redcross & \redcross & \redcross & \greencheck \\ 
        \midrule
        
        \multirow{4}{*}{Auto Gen} & Controllable Layout & \greencheck & \redcross & \redcross & \redcross & \redcross & \greencheck & \redcross & \redcross & \redcross & \redcross & \redcross \\
        & AI-generated Task & \greencheck & \greencheck & \redcross & \redcross & \greencheck & \redcross & \redcross & \redcross & \redcross & \redcross & \redcross \\
        & Machine-generated Data & \greencheck & \greencheck & \greencheck & \greencheck & \redcross & \redcross & \greencheck & \greencheck & \greencheck & \greencheck & \greencheck \\ 
        
         & Photorealism & RT & RT & R &  R & R & R & RT &  R & R &  R &  RT & \\ \bottomrule
        \end{tabular}
    }
    \label{tab:benchmark_comparison_transposed}
\end{table*}

Furthermore, we introduce \textbf{\benchmark}, a comprehensive benchmark built upon the scenarios generated by \method. As summarized in \cref{tab:benchmark_comparison_transposed}, \benchmark offers controllable layouts and multi-dimensional evaluation criteria, assessing policy generalization across object appearances, common-sense knowledge, spatial relationships, and long-horizon task completion. To support thorough benchmarking, \benchmark features 200 human-curated realistic scenarios and is designed to be scalable and adaptable, accommodating diverse robotic research directions. 

Based on \benchmark, we systematically evaluate existing robotic manipulation methods through:
\begin{itemize}
    \item \textbf{Modular manipulation.} 
    For zero-shot evaluation of visual prompting-based manipulation, we unify existing modular systems and diagnose errors in each module. Results indicate current prompt-based methods achieve certain zero-shot robustness (23\% SR), yet show clear limitations in long horizon tasks (11\% SR).
    \item \textbf{End-to-end manipulation.} 
    Leveraging large-scale demonstrations in \method, we investigate the generalization capabilities of end-to-end policies across diverse scenarios. Empirical results reveal that existing end-to-end approaches exhibit significant limitations in instruction-following and scenario generalization under limited pre-training data. These findings underscore the potential of the dual-system framework~\cite{figure_ai_helix} as a more scalable and robust alternative.
\end{itemize}




\section{Related Works}
\label{sec:relate}

\noindent\textbf{Simulation platforms for robotics.} Significant advancements have been made in developing robot simulators~\cite{rlbench, Ai2-thor, isaac}. Many simulation frameworks~\cite{mees2022calvin,zheng2022vlmbench,shridhar2020alfred,zeng2021transporter,gong2023arnold,jiang2022vima,pumacay2024colosseum,GemBench} have been built for robotics based on these simulators. But very few of them can stratify the requirement of diversity in the type of scenes, objects and activities, and realism of the underlying simulation environments as mentioned by Behavior-1K~\cite{behavior1k}. Moreover, generating demonstrations is crucial in the field of robot learning. Many methods have been developed to generate demonstration, including rule-based finite state machine~\cite{furniturebench} and transfering human demonstration~\cite{mimicgen,robocasa}. Differently, \benchmark incorporates hand-designed skills such as motion planner and grasping~\cite{anygrasp} by leveraging privileged information to generate feasible demonstrations.

\noindent\textbf{Task and scenario generation.} Scenario generation is widely used in robotics~\cite{robocasa,behavior1k,gensim,ProcTHOR} and autonomous driving~\cite{Scenic,realgen,driving_scenario_survey} to create diverse test environments that efficiently evaluate an agent's performance under various conditions, as real-world testing is often expensive and time-consuming. Prior benchmarks~\cite{mees2022calvin,simpleenv,GemBench} for tabletop tasks overlook the importance of well-designed scenarios, particularly in the era of MLLMs, where basic scenarios fail to sufficiently test the capabilities of advanced methods or provide comprehensive evaluations for multi-step processes. While basic domain randomization~\cite{domain_randomization} fails to produce challenging and realistic scenarios, manually crafted procedural generation processes~\cite{furniturebench} demand substantial expertise and labor. To tackle this issue, the scene graph is often utilized~\cite{Active_task_randomization}, where nodes represent objects and edges denote relations between these objects. This method captures the complex relationships within the scene~\cite{Visual_genome} and helps generate new data~\cite{Sceneverse,grutopia}. We present the \textit{task-oriented scene graph}-based scenario generation pipeline in our proposed \benchmark, which enables efficient and controlled, task-aligned scenario generation and assessment using scene graphs.

\noindent\textbf{Manipulation methods.} \textit{End-to-End imitation learning-based policies}~\cite{RT-1, jiang2022vima, MOO, cliport, Mobile_aloha, octo, aloha, perceiver-actor, goyal2024rvt, tian2024seer} rely on training policies with large demonstration datasets. Recently, several studies have explored fine-tuning large pre-trained VLMs~\cite{RT-1, openxembodiment, RT-2, embodiedgeneralist, roboflamingo} to enhance policy generalization across new scenes and tasks. These approaches require a large volume of demonstration trajectories due to limited inductive bias. Intermediate guidance techniques such as grounding~\cite{MOO}, trajectory~\cite{rt-trajectory, wen2023atm, robotap}, key-points~\cite{qin2020keto}, flow~\cite{im2flow2act, eisner2022flowbot3d, seita2023toolflownet, yuan2024generalflow, track2act}, and point clouds~\cite{wang2022goal, 3ddiffusionpolicy} have been recently introduced to mitigate this limitation. Modular Manipulation works Leveraging large multi-modal models like GPT-4V/GPT-4o~\cite{GPT-4}, recent works~\cite{voxposer, copa, moka, Manipulate-anything, vila, codeaspolicy} employing foundation models~\cite{sam, groundedsam, sam2} have demonstrated promising generalization capabilities in custom-designed scenarios. However, these designs are typically validated in self-designed tabletop settings and lack robust comparisons or benchmarks across diverse scenarios. To enable a fairer comparison between the two primary approaches, our main contribution is task and scenario simulation across diverse manipulation scenarios in simulation. This allows for a quantitative analysis of existing manipulation methods.

\section{\method}
\label{sec:method}

In this section, we introduce the overall simulation platform called \method. \cref{sec: simulation setups} describes the basic asset collections and the preparation of primitive skills, while \cref{sec: ToSG} outlines the complete pipeline for task and scenario layout generation.

\begin{figure*}[ht]
    \centering
    \includegraphics[width=1\linewidth]{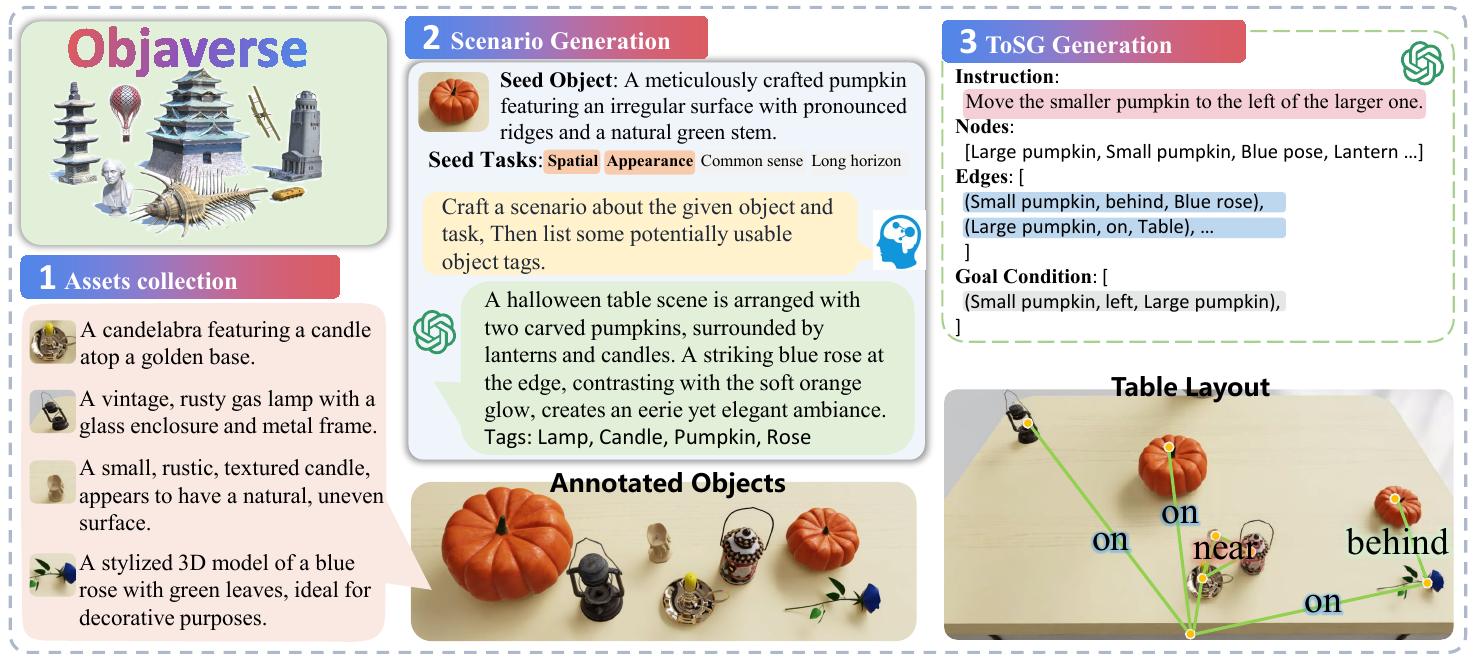}
    \caption{\textbf{Task-oriented scene graph (ToSG) for scenario synthesis.} We curate and annotate high-quality assets with GPT-4V. Given tagged objects, GPT-4 generates a \textit{task-oriented scene graph} (ToSG) with object states and relations. The ToSG guides layout construction and ensures relational consistency. Arrows between target and anchor objects are omitted for clarity. Goal conditions are used for evaluation.}
    \label{fig:layout gen}
\end{figure*}

\subsection{Simulation Platform Setups}\label{sec: simulation setups}

\method is built upon NVIDIA’s IsaacSim~\cite{isaac}, leveraging its photorealistic rendering and efficient parallel data collection capabilities. 

\noindent\textbf{Rigid assets with VL annotations.} Robust policy generalization across diverse manipulation tasks requires a large-scale, diverse, and richly annotated 3D object set. Thus, we curate high-quality assets by sourcing rigid objects from Objaverse~\cite{objaverse} and articulated ones from GRUtopia~\cite{grutopia} and PartNet-Mobility~\cite{sapien}, as shown in \cref{fig:teaser}. Our annotation pipeline includes:
\begin{enumerate}
\item \textbf{Filtering:} We reduce Objaverse from 660K to 70K assets by filtering based on tags and captions. We further exclude those with fewer than 1K mesh faces or missing normal textures.
\item \textbf{VL Annotation via GPT-4V:} For each asset, we generate structured annotations, including object description, physical properties (scale, mass), and semantic properties (category, color, shape, material).
\item \textbf{Centering and resizing:} We center and rescale 10K selected assets to tabletop size using GPT-4V, followed by human-in-the-loop corrections.
\end{enumerate}

\noindent\textbf{Articulated Assets.} We manually annotate 100 articulated objects (\eg, laptops, trash cans, ovens, microwaves, drawers), explicitly specifying their physical constraints.

\noindent\textbf{Diverse backgrounds.} To ensure environmental variety, we incorporate 100 indoor HDRI images, enriching the scenes with diverse backgrounds and lighting conditions.

\noindent\textbf{Primitive skills.} For pick-and-place tasks, we manually apply primitive skills such as \textit{pick}~\cite{anygrasp} followed by \textit{place}. We define key poses: pre-\textit{grasp}, post-\textit{grasp}, \textit{move}, pre-\{\textit{place}\}, and post-\{\textit{place}\}, where grasp and place poses are determined based on object height. A success check ensures task validity during scenario generation. For articulated objects, we follow MimicGen~\cite{mimicgen, skillmimicgen} and provide teleoperated trajectories based on object functionality (\eg, \textit{open}, \textit{close}, \textit{press}, \textit{push}, \textit{pull}, \textit{twist}). Scaling more diverse skill types remains an open challenge for future work.

\subsection{Task-oriented Scene Graph for Scenario Synthesis}\label{sec: ToSG}
To automatically synthesize diverse task scenarios from large-scale annotated assets, we represent each scenario in a structured yet intuitive format suitable for manipulation by LLMs. Traditional representations, such as PDDL~\cite{pddl}, explicitly define world states, initial conditions, actions, and goals. However, their rigid syntax, involving detailed action preconditions and state transitions, is cumbersome for LLMs, especially in complex or large-scale tasks. Thus, to simplify the intermediate representation, we introduce the \textit{task-oriented scene graph} (ToSG), a structured yet natural format explicitly designed for robotic manipulation scenarios involving spatial and relational reasoning. A ToSG consists of the following components:

\begin{itemize}
    \item \textbf{Task instructions}: a textual instruction designed to clarify and disambiguate the designed scene graph.
    \item \textbf{Scene graph}: nodes represent intra-object state (\eg, \textit{open}, \textit{closed}), denoted by:
        \[(\textit{object}, \textit{status}).\]
    and edges define inter-object relations, specified as:
        \[(\textit{object}, \textit{relation}, \textit{anchor object}).\]
        Possible relations include \textit{left}, \textit{right}, \textit{front}, \textit{behind}, \textit{near}, \textit{on}, \textit{in}, all relative to the camera viewpoint.
    \item \textbf{Goal conditions}: Defined as a list of condition items, each condition comprises either pairs (goal object state) or triplets (goal relationships).
\end{itemize}

\noindent\textbf{LLM-driven ToSG generation.} As illustrated in~\cref{fig:layout gen}, to ensure diverse yet coherent tasks, we first randomly sample initial seed objects using their visual-language (VL) annotations. Seed task types are selected across both short-horizon reasoning (spatial, appearance, and common sense) and long-horizon tasks. Relevant objects are retrieved from the annotated asset collection, typically sampled by tag, resulting in around 50 unique objects per scenario. GPT-4 then synthesizes task scenarios using detailed captions, outputting them in the \textit{task-oriented scene graph} (ToSG) format. This process promotes task diversity via varied object selection, while the structured ToSG ensures spatial and relational consistency. In contrast, RoboCasa~\cite{robocasa} focuses on skill-level task generation and limited scenes are pre-defined in certain randomization.

\begin{figure}
    \centering
    \includegraphics[width=1\linewidth]{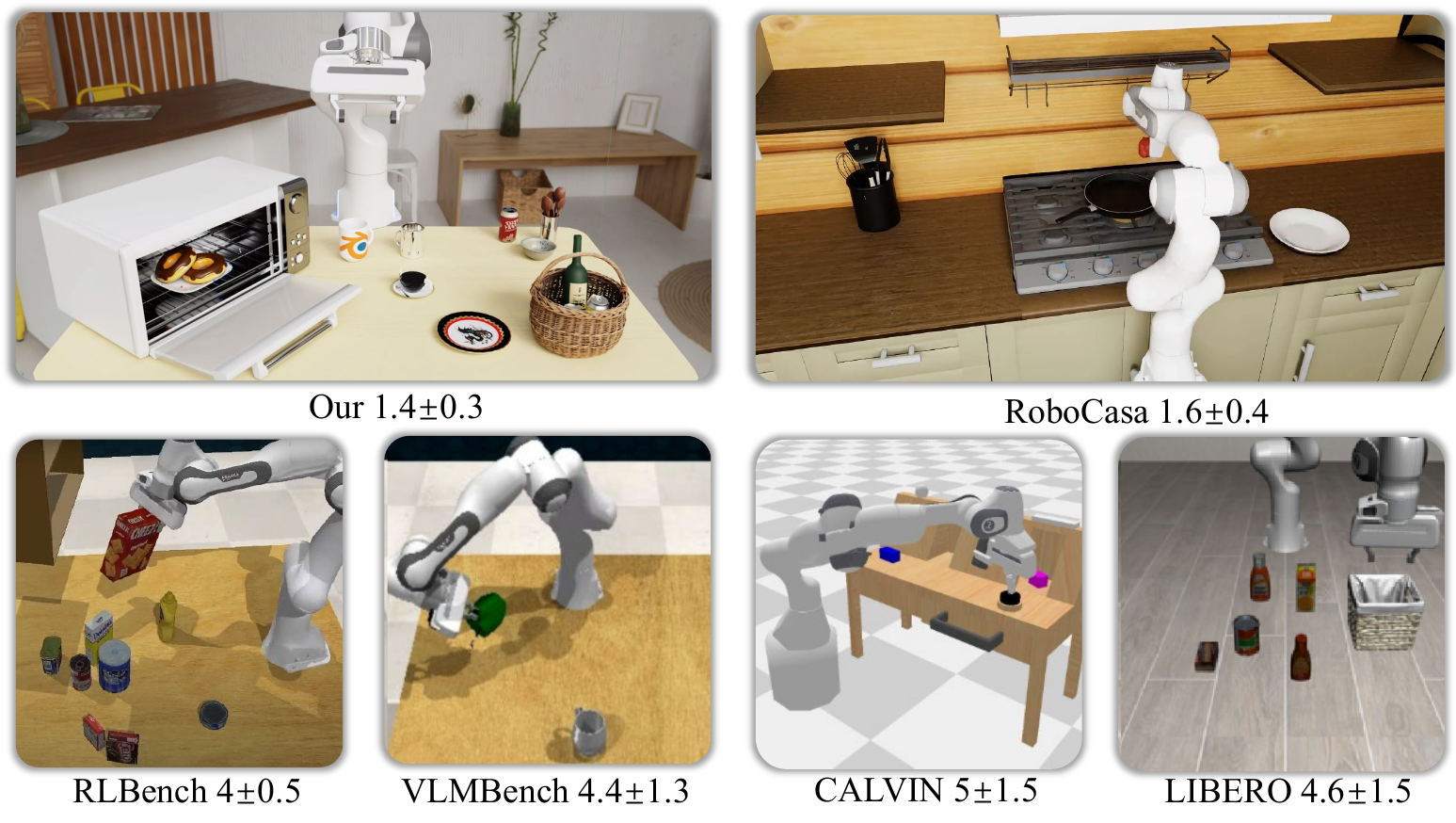}
    \caption{\textbf{User study.} We conduct a user study that encompasses popular robot benchmarks. We asked participants to rank the realism of 20 sampled scenarios from each environment on a scale of 1 (most realistic) to 5 (least realistic). We report the mean and standard deviation of the scores and provide a sampled image from the study.}
    \label{fig:user_study}
\end{figure}

\noindent\textbf{ToSG-based layout construction.} Based on the ToSG, we first sort all objects in a topological order using the \textit{on} and \textit{in} relations—for example, a table is placed before the objects on top of it. Next, objects at the same topological level are sequentially placed, resolving collisions while satisfying relational constraints from the scene graph. For instance, the relation "near" is defined within a specific threshold (e.g., 5 centimeters). Placements that violate the scene graph constraints are discarded. We ensure task executability by collecting at least one valid demonstration. See supplementary materials for details. A user study is conducted using standard robot benchmarks, as shown in~\cref{fig:user_study}.


\noindent\textbf{Automatic behavior-cloning data collection.} To study the generalization of learning-based manipulation, we provide automated data collection scripts. Leveraging privileged information about scene layouts and primitive skills—similar to MimicGen~\cite{mimicgen} (see \cref{sec: simulation setups})—we generate one demonstration per goal condition using a motion planning library~\cite{MPlib}.

\section{\benchmark: Benchmarking Instruction-Following Manipulation}
To systematically assess how manipulation policies generalize under diverse instructions and scenarios—a capability crucial for foundation model-based systems yet underexplored in current platforms—we present \benchmark, a curated set of 200 task scenarios refined through human-in-the-loop corrections (see supplementary for details). Based on GPT-synthesized scenes, human annotators refined each task to ensure diversity across both short-horizon reasoning and long-horizon goals, while preserving realistic layouts and everyday activities. 

\begin{figure*}[ht]
    \centering
    \includegraphics[width=1\linewidth]{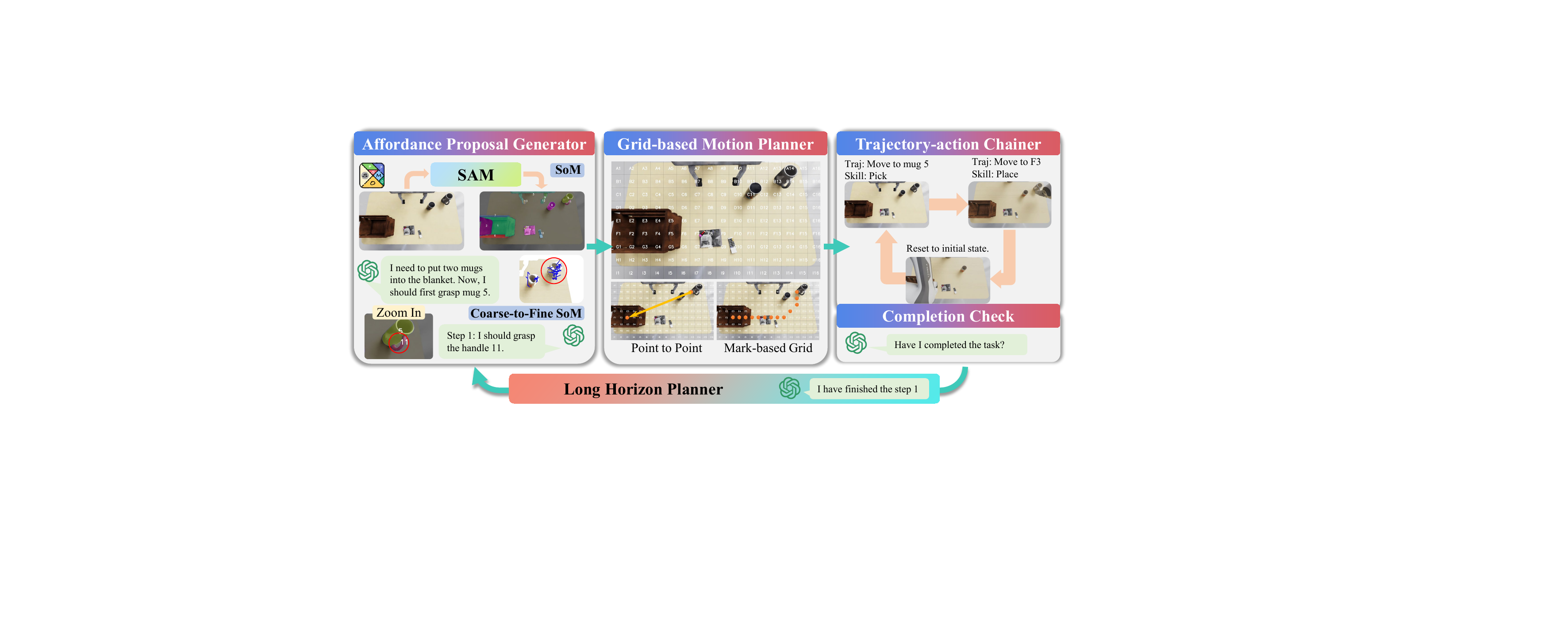}
    \caption{\textbf{Modular manipulation system via mark-based visual prompting.} The system includes four modules: (1)~\textit{Affordance proposal generator} extracts object masks using SAM2 and selects targets via SoM and coarse-to-fine SoM; filtered grasp poses are obtained with AnyGrasp. (2)~\textit{Grid-based motion planner} prompts the VLM to generate grid sequences with height and orientation, forming 3D trajectory points. (3)~\textit{Trajectory-action chainer} maps trajectory segments to high-level actions. (4)~\textit{Long-horizon planner} monitors task progress and triggers replanning if needed.}
    \label{fig:modular manipulation system}
\end{figure*}

\subsection{ToSG-based Evaluation}\label{sec: ToSG-based eval}
The policy executes for up to 9000 physics simulation steps in Isaac Sim, corresponding to approximately 150 seconds at 60 FPS. After execution, the final scene is evaluated to determine whether the goal conditions are satisfied.

\noindent\textbf{Evaluation metrics.} We evaluate the results using two widely adopted metrics: success rate (SR) and success rate weighted by path length (SPL). 
\begin{itemize}
    \item \textbf{SR:} Success is defined as the agent correctly meeting the goal conditions. For $M$ goal conditions, achieving one condition yields a score of $1/M$.
    \begin{equation}
        \text{SR Score} = \frac{1}{N}\sum_{i=1}^N \frac{1}{M}\sum_{j=1}^M1{\{\text{goal condition}\}}\
    \end{equation}
    \item \textbf{SPL:} SPL score is commonly used in visual navigation tasks~\cite{SPL} and calculated as follows:
    \begin{equation}
        \text{SPL Score} = \frac{1}{N}\sum_{i=1}^{N}S_i\frac{l_i}{\max(p_i,l_i)}
    \end{equation}
    where $N$ is the number of test episodes, $l_i$ is the shortest path distance from the initial position to the target position, and $p_i$ is the length of the path actually taken. $S_i$ denotes the success rate in the episode. 
\end{itemize}

\subsection{Modular Manipulation System}
Recent modular methods~\cite{moka, copa, google2024pivot, robopoint} leverage mark-based visual prompting in LLMs or foundation models to achieve strong generalization in real-world tasks. We abstract these strategies into a unified modular system, as shown in~\cref{fig:modular manipulation system}, to study the contributions of individual components. This framework includes four modules: an affordance proposal generator, a grid-based motion planner, a trajectory-action chainer, and a long-horizon planner.

\noindent\textbf{Affordance proposal generator.} We utilize SAM2~\cite{sam2} to generate masks for observations and label each mask with a unique number. The VLM then selects the target object’s mask based on instructions, a process known as Set-of-Mark (SoM) prompting~\cite{SoM}. Inspired by CoPA~\cite{copa}, we abstract it as multiple SoM calls to determine the precise location of the graspable item (\eg, mug handle), referred to as Coarse-to-Fine SoM (CtoF SoM). We employ AnyGrasp~\cite{anygrasp} to process RGB-D images, generating grasp pose proposals for the entire image. These proposals are then filtered using masks selected by SoM to obtain the grasp pose for the target object.

\noindent\textbf{Grid-based motion planner.} We adopt a strategy similar to MOKA~\cite{moka} for motion planning using VLMs. Specifically, we divide the image into $9\times 16$ grids and inform the VLM of the target object grid ID (the grid containing the mask center selected by the previous step) and instructions. Following MOKA, in a mark-based grid prompting setting, the VLM sequentially selects a series of grids along with the corresponding height above the table and gripper orientation. We term selecting a single target grid as point-to-point (choose only the ending point), a simplified version of constraint generation in CoPA~\cite{copa}. By reprojecting the selected grid centers along with the VLM's output height to 3D space, we obtain a series of trajectory points, which are then used to solve a series of actions using MPlib~\cite{MPlib}.

\noindent\textbf{Trajectory-action chainer.} Within the system, each sub-task can be divided into trajectory-action pairs. The term ``trajectory'' denotes the wide-range movement of the end effector, while ``action'' refers to script skills (\eg, \textit{pick}, \textit{place}). The agent must label the operation performed by the grid ID on the complete trajectory that was planned in the previous step. We leave the aggregation of diverse primitive skills via affordance prediction~\cite{geng2023gapartnet} for future work.


\begin{table*}[ht]
\centering
\caption{\textbf{Performance comparison of modular manipulation systems.} CoPA $^\dag$ and MOKA $^\dag$ are reproduced within \method with adaptations.}
\label{tab: main results}
\resizebox{0.88\textwidth}{!}{%
\begin{tabular}{lccccccccccc}
\toprule
\multirow{2}{*}{Modular methods} & \multirow{2}{*}{MLLM} & \multicolumn{2}{c}{Spatial} & \multicolumn{2}{c}{Appearance} & \multicolumn{2}{c}{Common Sense} & \multicolumn{2}{c}{Long-Horizon} & \multicolumn{2}{c}{Overall} \\ 
\cmidrule(lr){3-4} \cmidrule(lr){5-6} \cmidrule(lr){7-8} \cmidrule(lr){9-10} \cmidrule(lr){11-12}
& & SR & SPL & SR & SPL & SR & SPL & SR & SPL & SR & SPL \\ \midrule


\multirow{5}{*}{CoPA$^\dag$~\cite{copa}}    & GPT-4o & \textbf{25.00} & 4.57 & \textbf{29.41} & \textbf{7.12} & 20.59 & 4.19 & \textbf{17.59} & \textbf{4.26} & 20.33 & 4.53 \\
                                            & GPT-4.5 & \textbf{25.00} & \textbf{5.44} & 26.47 & 4.53 & \textbf{29.41} & \textbf{7.64} & 11.11 & 2.97 & \textbf{23.00} & \textbf{5.26} \\
                                            & Claude-3.7-Sonnet & 8.33 & 0.59 & 14.71 & 2.15 & 2.94 & 0.26 & 7.41 & 1.08 & 8.67 & 1.15 \\
                                            & Gemini-2.0-Flash & 8.33 & 1.81 & 38.24 & 7.83 & 17.65 & 4.12 & 8.33 & 1.76 & 19.00 & 4.02 \\
                                            & Qwen2.5-VL-72B & 8.33 & 2.01 & 23.53 & 5.60 & 10.78 & 1.96 & 12.96 & 2.72 & 13.67 & 3.05 \\
 \midrule
\multirow{5}{*}{MOKA$^\dag$~\cite{moka}}    & GPT-4o & 8.33 & 0.50 & 11.76 & 1.28 & 11.76 & 2.15 & 5.56 & 0.63 & 10.00 & 1.29 \\
                                            & GPT-4.5 & \textbf{16.67} & \textbf{3.75} & \textbf{23.53} & \textbf{2.20} & 17.65 & 3.26 & \textbf{11.11} & \textbf{2.25} & \textbf{14.00} & 2.14 \\
                                            & Claude-3.7-Sonnet & 0.00 & 0.00 & 0.00 & 0.00 & \textbf{20.59} & \textbf{6.02} & 8.33 & 1.12 & 7.00 & 2.05 \\
                                            & Gemini-2.0-Flash & 8.33 & 0.72 & 29.41 & 7.31 & 0.00 & 0.00 & 0.00 & 0.00 & 12.00 & \textbf{2.66} \\
                                            & Qwen2.5-VL-72B & 8.33 & 1.40 & 17.65 & 3.73 & 8.82 & 0.70 & 8.33 & 1.64 & 11.00 & 1.84\\
\bottomrule
\end{tabular}
}
\end{table*}


\noindent\textbf{Long horizon planner.} 
The long horizon loop encompasses task decomposition and completion checks. Task decomposition necessitates that agents analyze the instructions, historical context, and current situation to ascertain the appropriate subsequent step. The completion feedback evaluates whether the agent must revert to call the first module if the task is not yet completed. This module not only supports the execution of long horizon tasks but also provides the agent with a mechanism to recover from any failures in preceding operations.




\begin{table}[ht]
\caption{\textbf{Modular system ablation study. Effects of long-horizon tasks (left) and robustness to object distractors (right).} ``Horizons" refers to the number of goal conditions and ``Distractor" refers to the number of objects belonging to the same category.}
\resizebox{0.45\textwidth}{!}{
\begin{tabular}{cccccc}
\toprule
Horizons & SR (\%) & SPL (\%) & Type & \# & SR (\%)  \\ \cmidrule(lr){1-3}\cmidrule(lr){4-6}
1 & 32.28 & 10.12 & \multirow{3}{*}{\# Distractor} & 0 & 20.27\\
2 & 18.69 & 5.40 &  & 1 & 18.60\\
3 & 0.00 & 0.00 &  & 2 & 18.00\\
\bottomrule
\end{tabular}
}
\label{tab: ablation}
\end{table}


\subsection{Learning-based Manipulation}

For learning-based manipulation, we adopt two representative end-to-end policies: GR-1 and ACT. GR-1~\cite{gr1} is a GPT-style model that jointly processes language, vision, and robot states for action prediction. ACT~\cite{aloha, Mobile_aloha} is a transformer-based policy that improves robustness through visual augmentations and temporal ensembling.

\section{Experiments}
\label{sec:exp}

Our experiments aim to address two key questions: (1) What insights can a well-designed benchmark provide about current modular VLM agents? (2) Can learning-based methods generalize across varying task complexities and benefit from data scaling?



\subsection{Main Results}

We evaluate two prompt-based methods, MOKA~\cite{moka} and CoPA~\cite{copa}, within our benchmark framework, with detailed results in~\cref{tab: main results}. CoPA, leveraging GPT-4.5's capabilities, achieves a SR of 23.0\% across all tasks, significantly outperforming other models. This demonstrates CoPA's superior performance in handling diverse tasks.
However, the evaluation also highlights the challenges posed by long-horizon tasks, which emerge as the most difficult category. All models, including CoPA and MOKA, achieve an average success rate of 9.07\% on these tasks. This underscores a limitation of prompt-based models in managing the splitting of complex tasks and planning across a long horizon.

In contrast, for tasks that require understanding the appearance and everyday properties of objects, the models exhibit better performance. This suggests that the generalization capabilities of MLLMs enable them to comprehend and process the visual and functional attributes of objects, thereby allowing them to decompose and solve some long-horizon tasks more effectively.
A notable observation from the results is the success weighted by SPL metric. For CoPA, the SPL is approximately one-sixth of its task success rate, indicating a relatively efficient path planning performance. On the other hand, MOKA's SPL in some tasks is as low as one-tenth of its success rate, highlighting relatively inefficiencies in its path planning strategies.
These findings suggest that while prompt-based models like CoPA and MOKA can solve tasks to some extent, their reliance on intermediate representations rather than end-to-end outputs poses challenges for optimal path planning. This leads to suboptimal task execution paths, emphasizing the need for advancements in integrating end-to-end learning approaches to enhance the overall efficiency and effectiveness of robotic manipulation systems.

\begin{table}
\centering
\caption{\textbf{Ablation studies of modular manipulation system.} 
Following CoPA~\cite{copa}, we compare SoM prompting with AnyGrasp~\cite{anygrasp} for grasp proposal generation. ``CtoF SoM'' uses coarse-to-fine grounding. Point-to-Point plans motions from object centers to targets. ``Mark-based Grid'' from MOKA~\cite{moka} places grid points for planning. 
}
\label{tab: LLM ablation}
\resizebox{0.45\textwidth}{!}{%
\begin{tabular}{ccccccccc}
\toprule
Grasping Proposal & Motion Planning  & SR (\%) & SPL (\%) \\ \midrule
SoM & PointToPoint &  35.0 & 10.41\\ 
CtoF SoM & PointToPoint & 27.5 & 7.29\\
SoM & Mark-based Grid & 15.0 & 1.58\\
CtoF SoM &  Mark-based Grid & 10.0 & 0.94\\ 
\toprule
\rowcolor{gray!20} SoM & Oracle & 50.0 & -\\
\rowcolor{gray!20} CtoF SoM & Oracle & 65.0 & -\\
\bottomrule
\end{tabular}
}
\end{table}

\subsection{Ablation Studies for Modular Manipulation}

\noindent\textbf{Effect of distractors.} 
As shown in~\cref{tab: ablation}, the performance drops by 1.67 with the introduction of a single distractor but only decreases by an additional 0.6 with the more distractors.
This suggests a threshold effect, where the presence of initial distractors has a significant impact on performance, while further additions result in diminishing returns.

\noindent\textbf{Effect of task horizon.} 
As shown in~\cref{tab: ablation}, we select several tasks to evaluate the model’s performance on long-horizon tasks, categorizing them into different horizons.
The results show a significant decline in SR and SPL as the horizons increases.
Analyses indicate that these errors mainly stem from the task segmentation phase, where the model fails to decompose tasks correctly. 
This is largely due to the model’s inability to map everyday dialogue to task instructions or to correlate instructions with the corresponding images.

\noindent\textbf{Modular ablation.} 
Based on our pipeline design, we swap each module and evaluate their performance on a subset of our benchmark.
As shown in~\cref{tab: LLM ablation}, Point2Point motion planning significantly outperforms Mark-based Grid, indicating that GPT struggles with planning when required to handle continuous temporal sequences alongside image grounding. 
Additionally, CtoF SoM shows lower performance compared to SoM, as its grasping region analysis overlaps with AnyGrasp’s grasp pose proposals.
Despite AnyGrasp's providing reliable proposals, repeated SoM calls increase error likelihood and result in decreased performance.
\subsection{Ablation Studies for Learning-based Manipulation} 

\begin{figure}[th]
\centering
\includegraphics[width=1.\linewidth]{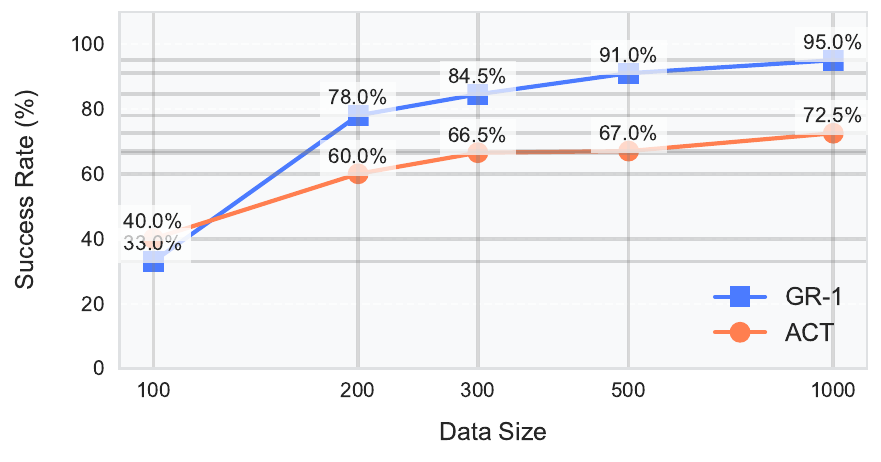}
\caption{\textbf{Ablation studies of data scaling effects.} }
\label{fig:scaling}
\end{figure}

We train two end-to-end learning-based policies, GR-1~\cite{gr1} and ACT, using varying data sizes from 100 to 1K training episodes, as summarized in~\cref{fig:scaling}. The evaluation settings range from constrained to highly diverse distributions: (1) \textit{Limited region randomization (fg)} randomizes only the target object within a $20\times 30$cm$^2$ area, with fixed distractors and instructions; (2) \textit{Limited region randomization (all)} randomizes all object positions within the same limited region; (3) \textit{Full table randomization (fg)} extends target object randomization to the entire table; (4) \textit{Full table randomization across 5 scenes} involves training and testing across different scene graphs and layouts; (5) \textit{Unseen instructions} modifies instructions in 5 scenes; and (6) \textit{Unseen objects} (scenarios) introduces object instances not seen during training.

\begin{figure}[th]
\centering
\includegraphics[width=1.\linewidth]{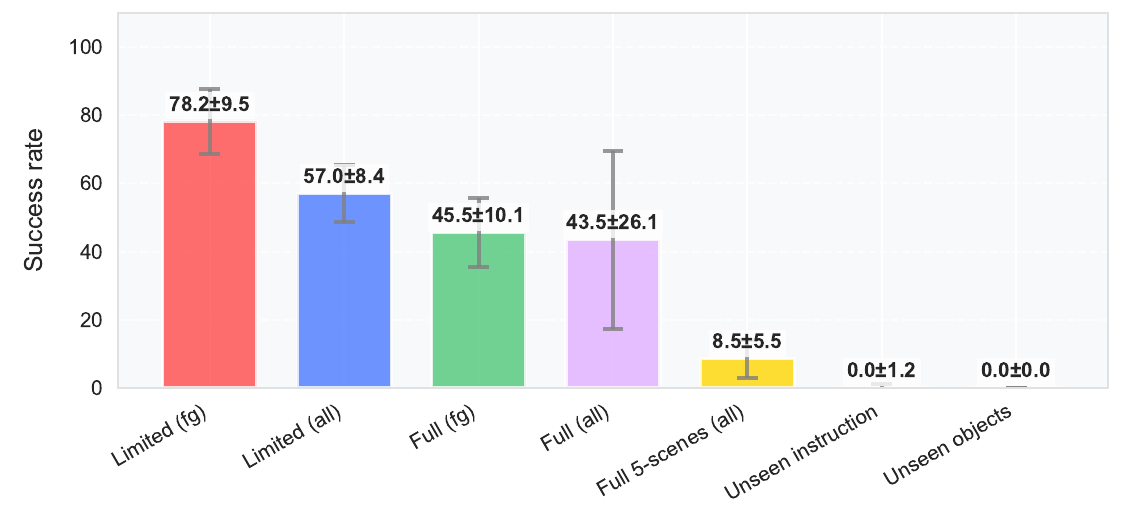}
\caption{\textbf{Ablation studies of end-to-end method generalization.} ``fg'' randomizes the target object's position; ``all'' randomizes all objects' positions. ``Limited'' denotes randomization on limited $ 20\times 30$ cm$^2$ area while ``Full'' denotes full table randomization.}
\label{fig:generalization}
\end{figure}

\noindent\textbf{Effects of BC data scaling.}
We evaluate GR-1 and ACT under a single-task setting—pick banana and place it on the plate—across varying data sizes. As shown in~\cref{fig:scaling}, both policies \textit{consistently} improve with more behavior cloning data. GR-1 demonstrates strong scaling, reaching 95.0\% SR with 1k episodes, up from 33.0\% at 100 episodes. ACT also improves steadily, from 40.0\% to 72.5\% SR over the same range. These results confirm that the data scale is valid for learning-based manipulation.

\noindent\textbf{Ablation study of model generalization.}
Despite strong performance under constrained conditions, GR-1 still exhibits limited generalization. As shown in~\cref{fig:generalization}, the model achieves 78.2\% SR in limited region randomization, but performance drops under broader randomization 45.5\% in full table randomization settings. Notably, in the same scene, position-level generalization is feasible, yet layout-level transfer remains challenging. Generalization across five scenes further reduces performance (43.5\% SR), and the model completely fails on Unseen instruction (0.0\% SR) and Unseen objects (0.0\% SR). These results suggest that end-to-end methods struggle to generalize beyond spatial variations to novel semantics and compositional scene changes. A full-scale evaluation across all 200 scenarios is left for future work due to computational constraints.

\section{Conclusion}\label{sec:conc}
We introduced \method, a large-scale simulation framework for benchmarking generalist robotic manipulation across diverse scenes and instructions. Central to it is the \textit{task-oriented scene graph} (ToSG), a scalable, natural language–based representation that models object-level and spatial relations for flexible task generation. We also present \benchmark, a set of 200 realistic scenarios for evaluating generalization. Experiments show that modular methods generalize better but struggle with spatial perception, while end-to-end policies benefit from scaling but remain sensitive to scene and instruction changes.

\section{Acknowledgements}
This work is funded in part by the National Key R\&D Program of China (2022ZD0160201), and Shanghai Artificial Intelligence Laboratory.

{
    \small
    \bibliographystyle{ieeenat_fullname}
    \bibliography{main}
}

\clearpage
\clearpage
\onecolumn
\setcounter{section}{0}

\renewcommand{\thesubsection}{\Alph{subsection}}

\begin{center}\Large\bf
Supplementary Materials for \method
\end{center}

The supplementary materials are organized as follows:

\begin{enumerate}[leftmargin=*]

    \item \textbf{Video demo:} We provide a supplementary video demonstration of \method, which includes additional examples of task scenario visualization and the real-world deployment of modular systems.
    
    \item \textbf{Experimental setups} are shown in \cref{sec: genmanip simulator and benchmark setups}.

    \item \textbf{\method task and scenario generation}:
    \begin{enumerate}
        \item Prompts for task-oriented scene graph generation are included in \cref{sec: Prompts of Task-oriented Scene Graph Generation}.

        \item Pseudo code for layout generation given a scene graph is shown in \cref{sec: layout generation based on scene graph}.
        
        \item The pipeline for human-in-the-loop corrections of benchmark scenarios is shown in \cref{sec: human-in-the-loop correction}.
        
        \item Example visualizations of task-oriented scene graphs and layouts are shown in \cref{sec: Task-oriented Scene Graph and Layout Visualization}.

    \end{enumerate}

    \item \textbf{\method demonstration collection:}
    \begin{enumerate}
        \item Articulated objects assets and primitive skills teleoperation are described in \cref{sec: articulated objects and their primitive skills}.
        
        \item Implementation details about BC data collection are included in \cref{sec: Implementation Details about Behavior-Cloning Data Collection}.
    \end{enumerate}

    \item \textbf{\benchmark:}
    \begin{enumerate}
        \item Implementation details about evaluating scene graph relations are shown in \cref{sec: evaluation}.
    
        \item Implementation details and prompts about modular manipulation systems, including visual prompts and step-by-step implementation, are included in \cref{sec: Implementation Details about Modular Framework}.    
        
        \item Implementation details of learning-based methods are shown in \cref{sec: Implementation Details about Learning-based Models}.
    
        \item Failure case visualizations of the modular systems are shown in \cref{sec: failure case visualization}.    
    \end{enumerate}
    
    \item Characterizing the sim-to-real gap is discussed in \cref{sec: characterizing sim-to-real gap}. 
\end{enumerate}

    

    

\section{Experimental Setups}\label{sec: genmanip simulator and benchmark setups}

\begin{wrapfigure}{r}{0.5\textwidth}
  \centering
  \includegraphics[width=0.5\textwidth]{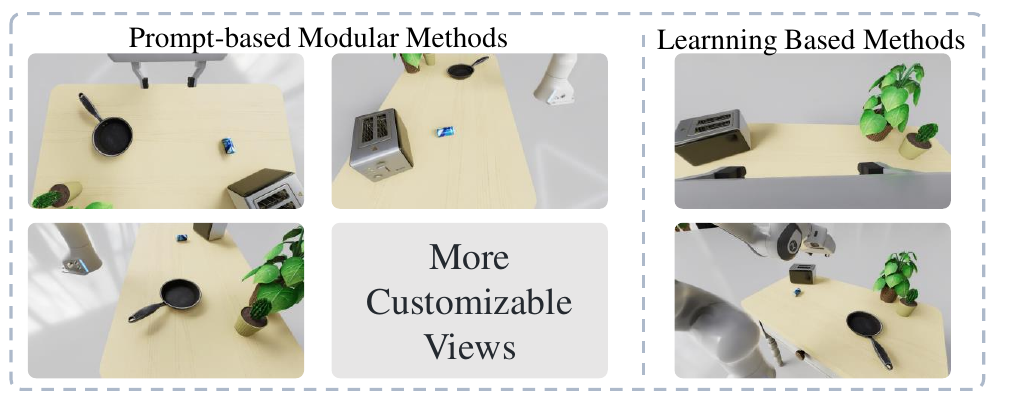}
  \caption{Camera setups for modular manipulation systems and learning-based methods in \benchmark.}
  \label{fig: camera views}
\end{wrapfigure}

\method operates within a tabletop scenario using a single Franka Arm on Isaac Sim 4.1.0. It features three distinct camera positions for prompt-based methods, along with two additional positions for data collection and testing of learning-based methods, as depicted in Figure A~\ref{fig: camera views}. By default, each camera captures RGB images and can optionally provide depth images and point clouds. We have designed user-friendly interfaces for easy adjustment of the cameras' intrinsic and extrinsic parameters. Users can also add or remove cameras to customize views and tailor benchmarks to their specific methods. In the \benchmark, we set the scene background to white to achieve a cleaner view and simplify the testing context.


We encapsulate potential model calls (e.g., Anygrasp~\cite{anygrasp}, SAM2~\cite{sam2}, custom learning-based models) as backend services, interacting with the main program. This architecture facilitates asynchronous communication between the models and the program, enhancing usability and reusability.

\section{Prompts of ToSG Generation} \label{sec: Prompts of Task-oriented Scene Graph Generation}
The base prompt for generating the task-oriented scene graph is shown in Prompt~\hyperref[prompt:ToSG]{1}. To generate four types of tasks, we incorporate task-specific prompts into the base prompt. Note that we do not include a long-horizon prompt, as this type of task can be derived from other tasks. The task specific prompts are listed in Prompt~\hyperref[prompt:Spatial_Reasoning]{2}, \hyperref[prompt Appearance Reasoning]{3} and \hyperref[prompt:ToSG]{4}.

\phantomsection
\label{prompt:ToSG}
\begin{tcolorbox}[colback=yellow!10!white, colframe=black!75!white, title=Prompt 1: ToSG Generation Prompt]
\scriptsize
You are an assistant specializing in simulation scene design. Your task is to select at least [Num\_of\_objects] objects from the provided list (each with specific names and descriptions) to include in a simulation scene. Each object should be strategically positioned relative to at least one other object using the specified spatial orientations: left, right, front, back, and top.\\

\textbf{Available Objects and corresponding states:}

- Name: Porcelain Plate

  UID: 008
  
  Description: A round, white porcelain plate with intricate blue floral patterns.
  
  States: None

// omitted

\textbf{Your Task:}

Design a task-oriented scene graph consisting of the following three parts:

1. \textbf{Instruction:} You need to design a tabletop manipulation task that will change the layout of the objects. If the task does not require changing the object's state, it should mimic a typical 'pick and place' activity as seen in daily life. However, if the task involves altering the object's state—such as opening or closing a cabinet—select an appropriate verb to accurately describe the action needed to modify the object's state.

   - Focus on clear and functional interaction between the objects.
   
   - When you need to refer to an object, you can use the provided name. You can change the name based on the instruction, but please do not create any ambiguity.

   - The goal condition is sufficient to judge the instruction.

   - [Task specified prompt]

2. \textbf{Goal Conditions:} Define the objectives of the Instruction, specifying:

   - The names and unique identifiers (UIDs) of the objects involved.
   
   - Their final states (choose from the given states; if none, state as 'none').
   
   - The relative positions intended between the objects (using 'front', 'back', 'left', 'right', 'near', 'top').

   - If multiple conditions are applied, they should be expressed in disjunctive normal form. Each atomic variable should determine whether a single spatial relation or object state is satisfied.

3. \textbf{Scene Graph:} Design a scene graph that captures the spatial relationships and states of the objects in the scene.

   - Begin by describing the initial scene, emphasizing the spatial relationships among objects within a natural setting. This will help you write the scene graph.
   
   - Each edge in the scene graph must have two objects.
   
   - Include sufficient edges to represent all connections and interactions.

\textbf{Note:} Assume all objects are on the table by default, so avoid specifying 'top' or spatial relations with it. The instruction must alter the initial layout, and the goal condition should not match the initial scene graph. Avoid the common mistake of creating a circular transformation where reversing the objects’ positions results in the same spatial relationships (e.g., "A on the left of B" becoming "B on the right of A").

\textbf{Output Format:}

Return the output in the following JSON format:
\renewcommand{\ttdefault}{cmtt}
\begin{verbatim}
{
    "instruction": "Instruction for the task", // within 30 words
    "goal_conditions": // disjunctive normal form: multiple conditions combined with AND inside [], connected by OR between []. 
    Example structure provided below.
    [
        [
        	{
            "obj1": "Name of the first object for the task",
            "obj1_uid": "UID of the first object",
            "obj1_state": "state of obj1",
            "obj2": "Name of the second object for the task, or 'none' if not needed",
            "obj2_uid": "UID of the second object, or 'none' if not needed",
            "position": "front/back/left/right/near/top, or 'none' if not needed"
        	}
        ]
    ],
    "scene_graph": {
        "description": "Describe the initial scene layout, emphasizing the spatial relationships between objects.
        Include details about the objects involved and their specific positions relative to one another.",
        "edges": [
            {
                "obj1": "Name of the first object",
                "obj1_uid": "UID of the first object",
                "position": "front/back/left/right/near/top",
                "obj2": "Name of the second object", // can not be None
                "obj2_uid": "UID of the second object"
            }
            // ...
        ],
        "nodes": [
            {
                "obj_uid": "UID",
                "state": "initial state" 
            }
            // ...
        ]
    }
}
\end{verbatim}
\renewcommand{\ttdefault}{pcr}

\end{tcolorbox}

\phantomsection
\label{prompt:Spatial_Reasoning}
\begin{tcolorbox}[colback=yellow!10!white, colframe=black!75!white, title=Prompt 2: Spatial Reasoning Task Prompt]
Design an instruction that assesses a model's spatial reasoning ability, which is the capability to understand, interpret, and infer indirect spatial relationships among objects to determine their precise locations. The instruction must clearly describe a task that involves interpreting complex spatial configurations. 

\end{tcolorbox}

\phantomsection
\label{prompt Appearance Reasoning}
\begin{tcolorbox}[colback=yellow!10!white, colframe=black!75!white, title=Prompt 3: Appearance Reasoning Task Prompt]
Design an instruction that evaluates a model's ability to reason about and recognize visual attributes such as color, size, shape, material, and other distinctive characteristics. The instruction must use these visual traits to uniquely identify objects without relying on their names or identifiers.
\end{tcolorbox}

\phantomsection
\label{prompt Common Sense Reasoning}
\begin{tcolorbox}[colback=yellow!10!white, colframe=black!75!white, title=Prompt 4: Common Sense Reasoning Task Prompt]
Design an instruction to evaluate a model's common sense reasoning. Present a clear, everyday scenario that involves a practical problem or goal related to human needs. The model should apply basic principles like cause and effect or logical outcomes to choose one specific action that solves the problem. 
\end{tcolorbox}

\section{Pseudo code of Layout Generation based on Scene Graph}\label{sec: layout generation based on scene graph}

The pseudo code for the layout generation pipeline is presented in~\cref{alg:layout_gen}. 

\begin{algorithm}[ht]
\caption{\textbf{Layout construction pipeline.}}
\label{alg:layout_gen}

\begin{algorithmic}[1]
\STATE \textbf{Input:} Scene Graph $G$ from task-oriented scene graph
\STATE \textbf{Output:} Generated Layout $L$

\STATE Initialize $L \gets \emptyset$
\STATE $O \gets \mathrm{TopologicalSort}(G)$ 
\STATE $C \gets \mathrm{ExtractRelationalConstraints}(G)$

\FOR{$o_i$ in $O$}
    \STATE $T_i \gets \mathrm{GetTopologicalLevel}(o_i, G)$
    \WHILE{True}
        \STATE $P_i \gets \mathrm{FindPlacement}(o_i, T_i, L, C)$
        
        \IF{$\mathrm{ValidatePlacement}(P_i, o_i, C, L)$}
            \STATE $L \gets L \cup (o_i, P_i)$ 
            \STATE Break
        \ENDIF
    \ENDWHILE
\ENDFOR

\RETURN $L$

\STATE \textbf{Subroutine Definitions:}
\STATE $\mathrm{TopologicalSort}(G):$ Sort objects in $G$ based on `on' and `in' relations.
\STATE $\mathrm{ExtractRelationalConstraints}(G):$ Extract relational constraints such as "nearby" from $G$.
\STATE $\mathrm{GetTopologicalLevel}(o, G):$ Get the topological level of object $o$ from $G$.
\STATE $\mathrm{FindPlacement}(o, T, L, C):$ Calculate placement of $o$ based on level $T$, layout $L$, and constraints $C$.
\STATE $\mathrm{ValidatePlacement}(P, o, C, L):$ Validate placement $P$ of object $o$ against $C$ and $L$.
\end{algorithmic}
\end{algorithm}

\section{Human-in-the-Loop Corrections of \benchmark Scenarios} \label{sec: human-in-the-loop correction}

\begin{figure}[h]
    \centering
    \includegraphics[width=0.5\linewidth]{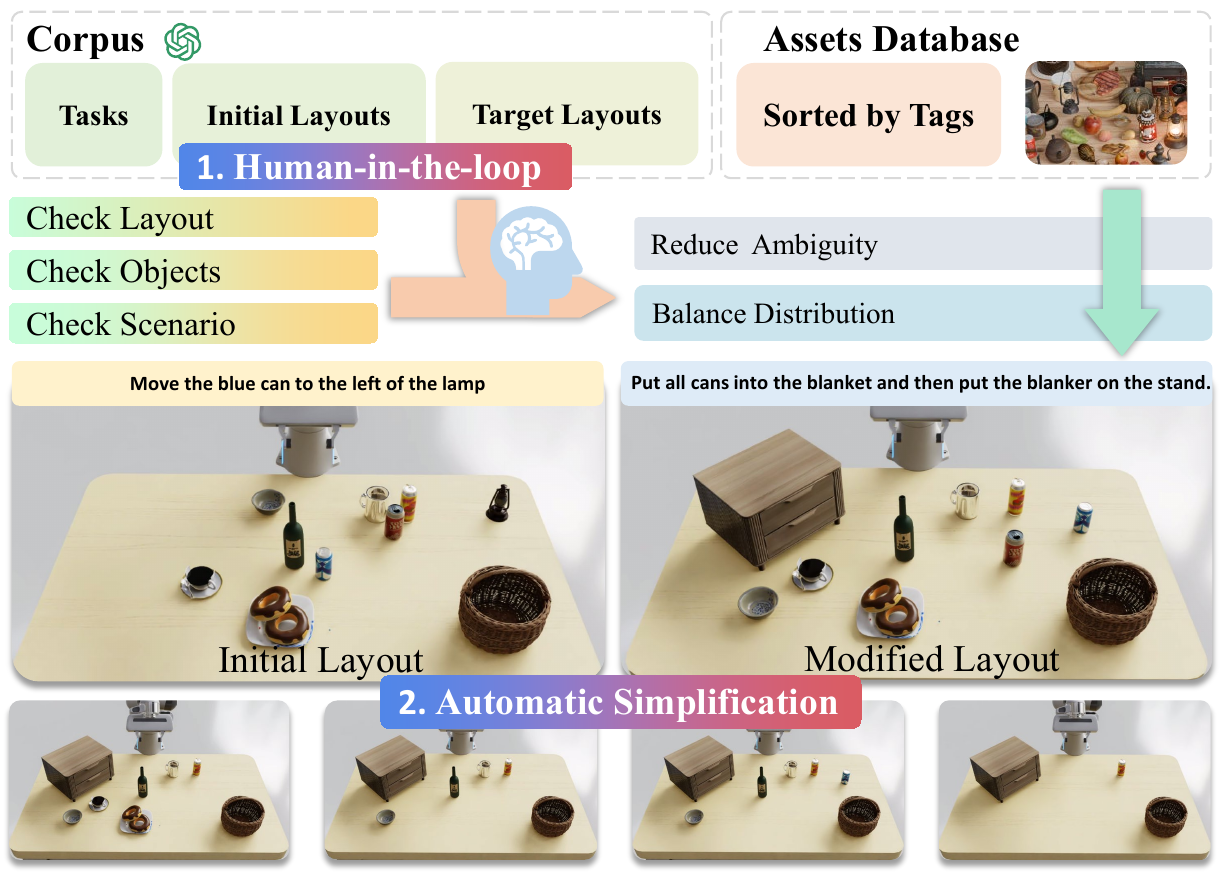}
    \caption{\textbf{Human-in-the-Loop corrections of benchmark scenarios.}}
    \label{fig: human in the loop}
\end{figure}

We begin by sampling four instruction types from the generated corpus to ensure comprehensive coverage of object appearances, common-sense knowledge, spatial relationships, and long-horizon tasks. Human annotators, with privileged access to the scene graph and USD files, use IsaacSim for detailed inspections. They also utilize a tagged, organized object collection for asset retrieval. Subsequently, annotators refine layouts, objects, and scenarios based on GPT-generated data to ensure daily consistency. They evaluate instruction feasibility, identifying potential issues such as collisions—for example, a robot grasping a blue can that interferes with a nearby bottle. Finally, to establish a balanced benchmark across generalization dimensions, we meticulously review the distribution of instructions and select 200 task scenarios for benchmarking.

\section{Example Visualization of ToSG and Layout} \label{sec: Task-oriented Scene Graph and Layout Visualization}

\begin{figure}[h]
    \centering
    \includegraphics[width=1\linewidth]{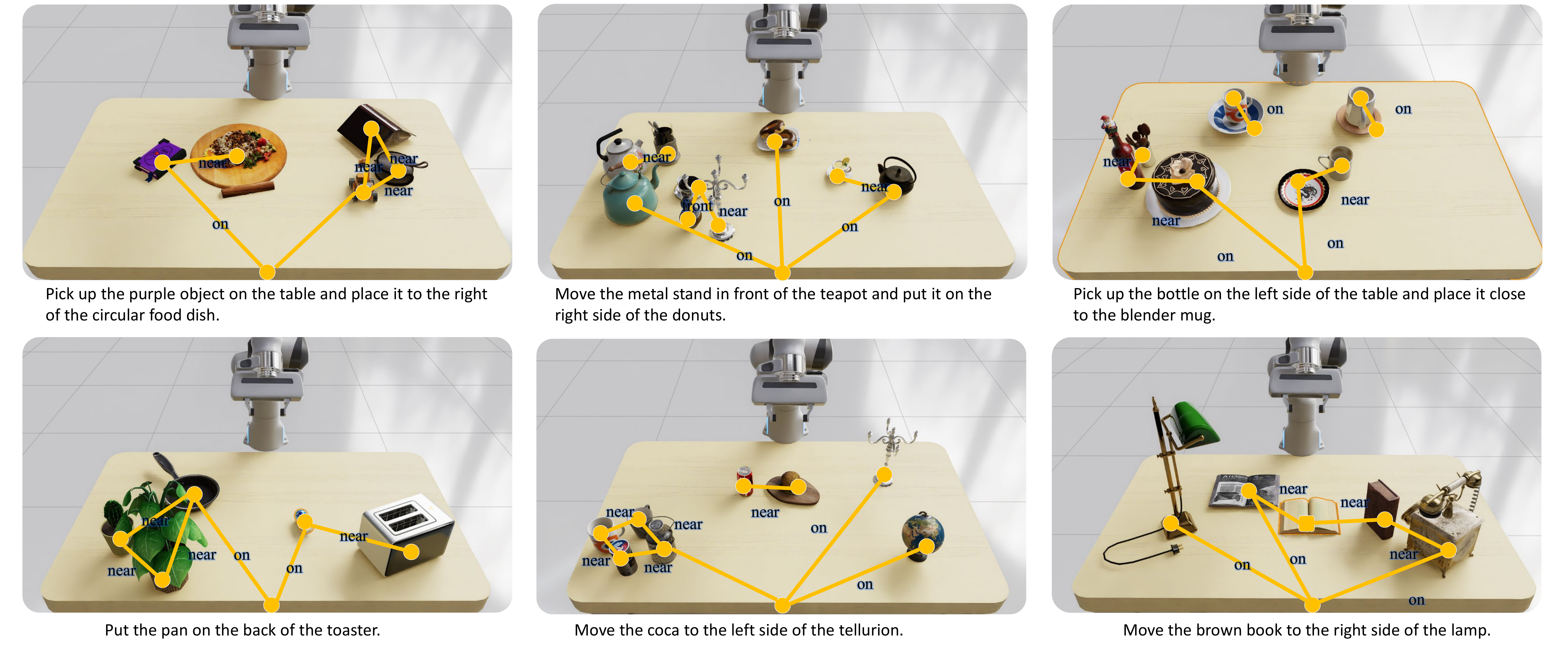}
    \caption{Example visualization of task-oriented scene graph and layouts.}
    \label{fig:scene_graph_visualization}
\end{figure}

We visualize example task-oriented scene graphs and their respective tabletop layouts in Figure A-~\ref{fig:scene_graph_visualization}.

\section{Articulated Objects And Teleoperation for Primitive Skills} \label{sec: articulated objects and their primitive skills}

\begin{figure}[h]
    \centering
    \includegraphics[width=1.0\linewidth]{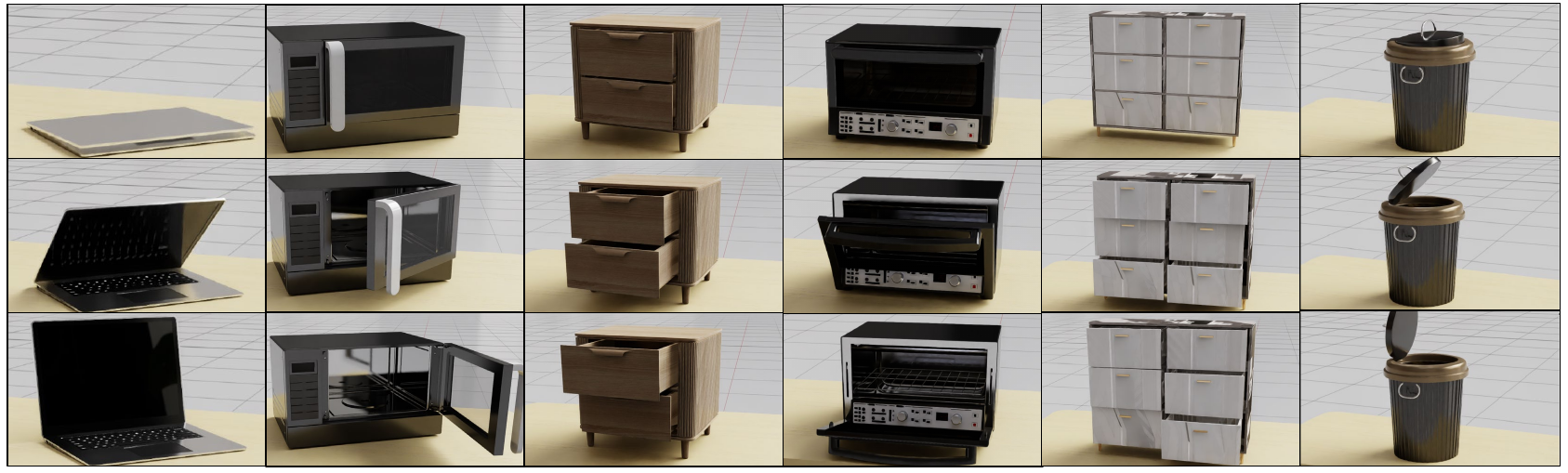}
    \caption{\textbf{Articulated objects.} The 6/100 articulated objects with different manipulation strategies are shown.}
    \label{fig:Articulated objects}
\end{figure}

\begin{figure}[h]
    \centering
    \includegraphics[width=1.0\linewidth]{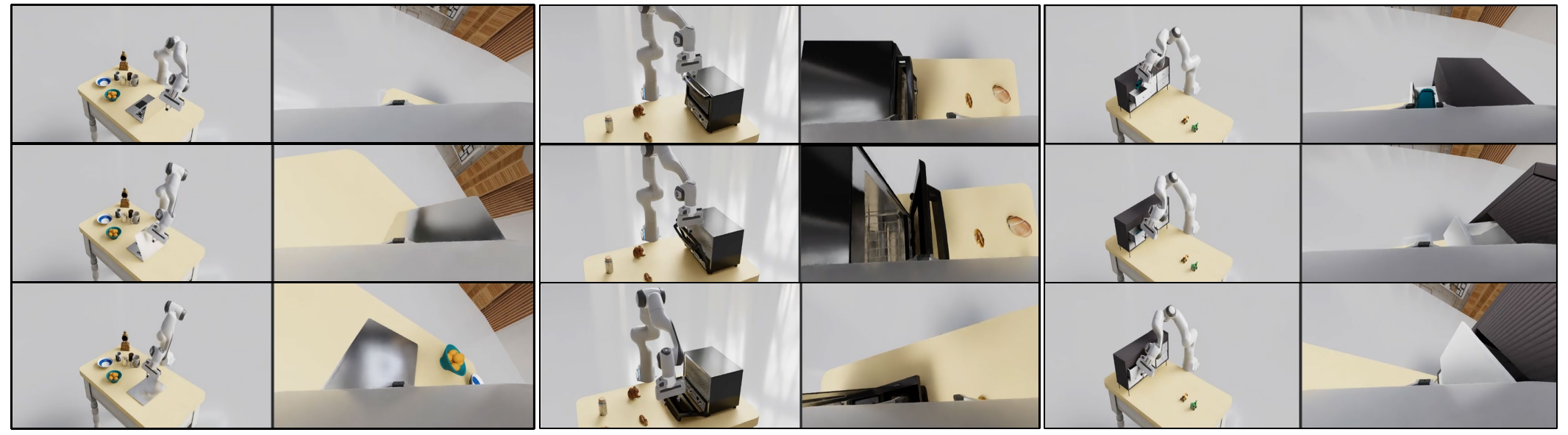}
    \caption{\textbf{Teleoperation.} Human annotators utilize the SpaceMouse, a 6-DoF device, to teleoperate the Franka robot in IsaacSim.}
    \label{fig:teleoperation}
\end{figure}


Figure A-\ref{fig:Articulated objects} displays several of these objects in Isaac Sim, which require specific manipulations. Following the approach of Mimicgen~\cite{mimicgen}, we collect primitive skills from human teleoperation trajectories for these articulated objects, as illustrated in Figure A-~\ref{fig:teleoperation}.

\section{Implementation Details about Behavior-Cloning Data Collection}\label{sec: Implementation Details about Behavior-Cloning Data Collection}

\begin{figure}[ht]
    \centering
    \includegraphics[width=1\linewidth]{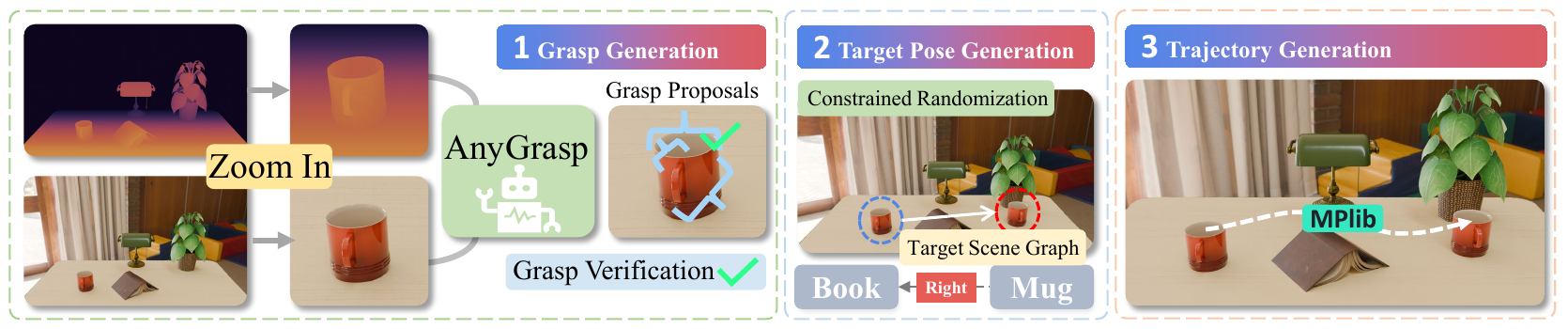}
    \caption{\textbf{BC data collection pipeline.} The pipeline comprises three steps: (1) grasp generation using AnyGrasp~\cite{anygrasp}, (2) target position generation constrained by the scene graph, and (3) trajectory generation with MPlib~\cite{MPlib}.
    }
    \label{fig:demonstration}
\end{figure}

The BC data collection pipeline is shown in Figure A-~\ref{fig:demonstration}. First, AnyGrasp~\cite{anygrasp} is utilized to generate the grasp pos of the end effector. To ensure high-quality grasp proposals, we crop the target object from the rendered RGB-D image and filter the grasp by direction to avoid potential collision between the end effector and the table. Then we randomly place the target object based on the goal scene graph. Finally, a trajectory is generated to connect smoothly the Franks initial pose, grasp pose, and target place pose with the help of MPlib~\cite{MPlib}. Each demonstration is tested in the physics-based simulator to ensure they can be implemented in the real world.

Specifically, for data collection, Isaac Sim is configured to render and perform physics calculations at 60 FPS, using the \texttt{plan\_pose} function from MPlib for motion planning, with \texttt{rrt\_range} set to 0.001. We use the PD Controller provided by Isaac Sim for force control of the Franka robot. Data collection involves recording RGB-D images from two camera perspectives. The initial resolution is $1280\times 720$, with depth in meters, and these images are resized to $512\times 288$. The color images are saved as JPG files, while the depth images, after being multiplied by 1000, are saved as PNG files. We choose the absolute joint positions as the representation of Franka's pose and record Franka's state and action at each frame. Additionally, we process the absolute joint positions into end effector coordinates in the robot's coordinate system and Euler angles in XYZ order as another form of representation.

\section{Evaluation Details} \label{sec: evaluation}

To evaluate object relationships, the final scenes post-method execution are converted into point clouds and the point clouds for each object are extracted. In this section, we detail the method for determining spatial relationships among these point clouds, encompassing horizontal, vertical, and multi-object interactions. The pseudo-code for assessing these spatial relationships is presented in~\cref{alg:spatial_relationship}.

\paragraph{Horizontal relations.}
Horizontal relations between two point clouds are evaluated based on their relative positions in the XY plane. The steps are as follows:

\begin{enumerate}
    \item Compute the 2D distance between the bounding boxes of the point clouds.
    \item If the distance is greater than a threshold \texttt{XY\_DISTANCE\_CLOSE\_THRESHOLD}, the point clouds are considered separate, and further horizontal relation analysis is skipped.
    \item If the distance is small enough, check for overlap along both the X and Y axes. The possible horizontal relationships are:
    \begin{itemize}
        \item \textbf{Left-Right}: This relationship is determined when the point clouds overlap along the X-axis but not the Y-axis. The objects are positioned side by side along the X-axis.
        \item \textbf{Front-Back}: This relationship is determined when the point clouds overlap along the Y-axis but not the X-axis. The objects are positioned side by side along the Y-axis.
        \item \textbf{Near}: If there is overlap in both the X and Y axes, the point clouds are considered to be near each other.
    \end{itemize}
\end{enumerate}

\paragraph{Vertical relations.}
Vertical relations between two point clouds are evaluated based on the Z-axis distance and their relative positions in 3D space. The key steps are:

\begin{enumerate}
    \item Calculate the vertical distance between the point clouds, including the distance from the top of one object to the bottom of the other.
    \item If the point clouds are close enough (within a threshold distance \texttt{MAX\_TO\_BE\_TOUCHING\_DISTANCE}), the function evaluates if one object is:
    \begin{itemize}
        \item \textbf{On/Beneath}: One object is on top of or below the other.
        \item \textbf{Supporting/Supported by}: If the objects are in contact or near each other, the function checks if one object supports the other based on the overlap area ratio.
    \end{itemize}
\end{enumerate}

\paragraph{Multi-Object Relations.}
When a third point cloud \texttt{point\_cloud\_c} is provided, the function computes the centroids of the point clouds and evaluates the angular relationship between the vectors formed by the centroids. The steps are as follows:

\begin{enumerate}
    \item Compute the centroid of each point cloud: \texttt{anchor1\_center}, \texttt{anchor2\_center}, and \texttt{target\_center}.
    \item Construct vectors \texttt{vector1} and \texttt{vector2} from the target centroid to each of the anchor centroids.
    \item Normalize the vectors and compute the cosine of the angle between them.
    \item If the angle is smaller than a predefined threshold \texttt{ANGLE\_THRESHOLD}, the relationship is labeled as \texttt{between}, indicating that the target object is positioned between the two anchor objects.
\end{enumerate}

\begin{algorithm}[h!]
\caption{Infer Spatial Relationship Between Point Clouds}
\label{alg:spatial_relationship}
\textbf{Input:} Point clouds $P_\mathrm A$, $P_\mathrm B$, (optional $P_\mathrm C$), bounding boxes $\min A$, $\max A$, $\min B$, $\max B$\\
\textbf{Output:} Spatial relationship between $A$ and $B$ (or $C$)\\
\textbf{Constants:} Constants for between, in/out of, above/below, on/beneath, near, left/right, front/back. 

\begin{algorithmic}[1]

\IF{$\mathrm P_C$ exists}
    \STATE $\vec{AB} \gets \mathrm{CalculatePointCloudVector}(\mathrm {GetCenter}(P_A), \mathrm {GetCenter}(P_B))$
    \STATE $\vec{BC} \gets \mathrm{CalculatePointCloudVector}(\mathrm {GetCenter}(P_B), \mathrm {GetCenter}(P_C))$
    \STATE Angle $\gets \mathrm{CalculateAngleByVectors}(\vec{AB}, \vec{BC})$ 
    \IF{ {IS\_BETWEEN}(Angle)}
        \RETURN \textbf{``between"}, \textbf{``between"}
    \ENDIF
\ELSE
    
    
    \STATE ${d}_{\mathrm{xy}} \gets \mathrm {CalculateXYDistance}(P_\mathrm A, P_\mathrm B)$
    \IF{IS\_INTOUCH(${d}_{\mathrm{xy}}$)}
        \IF{ IS\_INSIDE($P_\mathrm A$, $P_\mathrm B$)}
            \RETURN \textbf{``in''}, \textbf{``out of''}
        \ELSIF{IS\_INSIDE($P_\mathrm B$, $P_\mathrm A$)}
            \RETURN \textbf{``out of''}, \textbf{``in''}
        \ENDIF
        \IF{IS\_OVERLAP($P_\mathrm A$, $P_\mathrm B$, axis=z)}
            \IF{GetCenter($P_A$)[z] $>$ GetCenter($P_B$)[z]}
                \RETURN \textbf{``on"}, \textbf{``beneath''}
            \ELSIF{GetCenter($P_A$)[z] $>$ GetCenter($P_B$)[z]}
                \RETURN \textbf{``beneath''}, \textbf{``on''}
            \ENDIF
        \ELSIF{IS\_OVERLAP($P_\mathrm A$, $P_\mathrm B$, axis=x) and \textbf{not} IS\_OVERLAP($P_\mathrm A$, $P_\mathrm B$, axis=y)}
            \IF{GetCenter($P_A$)[x] $>$ GetCenter($P_B$)[x]}
                \RETURN \textbf{``front"}, \textbf{``back''}
            \ELSIF{GetCenter($P_A$)[x] $<$ GetCenter($P_B$)[x]}
                \RETURN \textbf{``back"}, \textbf{``front''}
            \ENDIF
        \ELSIF{IS\_OVERLAP($P_\mathrm A$, $P_\mathrm B$, axis=y) and \textbf{not} IS\_OVERLAP($P_\mathrm A$, $P_\mathrm B$, axis=y)}
            \IF{GetCenter($P_A$)[y] $>$ GetCenter($P_B$)[y]}
                \RETURN \textbf{``left"}, \textbf{``right''}
            \ELSIF{GetCenter($P_A$)[y] $<$ GetCenter($P_B$)[y]}
                \RETURN \textbf{``right"}, \textbf{``left''}
            \ENDIF
        \ENDIF
        \RETURN \textbf{``near"}, \textbf{``near''}
    \ENDIF
\ENDIF

\RETURN No Relationship

\STATE \textbf{Subroutine Definitions:}
\STATE GetCenter: Calculates the center point of the point cloud.
\STATE CalculatePointCloudVector: Obtains the vector through the center point of a point cloud.
\STATE CalculateAngleByVectors: Computes the angle between two vectors.
\STATE CalculateXYDistance: Determines the shortest distance between the projections of two point clouds on the XY plane.
\STATE IS\_BETWEEN/IS\_INTOUCH/IS\_INSIDE: Check if the input values satisfy the constraints defined by the respective relationships.
\STATE IS\_OVERLAP: Verifies whether the projection overlap of two point clouds along a specified coordinate axis meets the defined constraints.

\end{algorithmic}
\end{algorithm}

\newpage
\section{Implementation Details and Prompts about Modular Manipulation System} \label{sec: Implementation Details about Modular Framework}

In this section, we present the complete prompts of modular manipulation system. Specifically, we first use the prompts from the task decomposition by Prompt~\hyperref[prompt:SoM]{5} to divide the task, then use SoM by Prompt~\hyperref[prompt:SoM]{6} for mask selection. The CtoF SoM by Prompt~\hyperref[prompt:CtoF SoM]{7} is an optional component used to switch between different modules. After that, we run grab point selection by Prompt~\hyperref[prompt:Grab Point Selection]{8} to obtain the grab point and execute grid-based path planning by Prompt~\hyperref[prompt:Grid-based Path Planning]{9} or point-to-point path planning by Prompt~\hyperref[prompt:Point-to-Point Path Planning]{10} to acquire a series of waypoints. At this stage, we have obtained a series of two-dimensional coordinates in the image coordinate system, including the grab point and waypoints.

In the simulation evaluation with Isaac Sim, the prompt baseline acquires transformation-related information through the API for obtaining the camera's intrinsic and extrinsic parameters. It then uses reprojection to obtain a series of three-dimensional coordinates in the world coordinate system. Additionally, it calls Anygrasp~\cite{anygrasp} and uses the distance threshold between the grasp pose and the 3D grab point, along with score-based sorting, to determine the grasp pose.

Subsequently, the prompt baseline uses MPlib~\cite{MPlib} for motion planning, generating the joint positions from the grasp pose to each waypoint. It then uses a PD controller in Isaac Sim to perform force control, thereby controlling the Franka robot to execute the operation. After each execution of the modular methods, the Task completion checker by Prompt~\hyperref[prompt:Task completion checker]{11} is called to check the task completion status and achieve a closed-loop system.

Due to existing limitations in simulation skills, trajectory-action chaining currently only supports pick-and-place actions. However, the modular system retains its extensibility and can be equipped with more complex functionalities in the future. Additionally, the modular system is not coupled with Isaac Sim. In fact, the modular baseline only requires the camera's intrinsic and extrinsic parameters to obtain three-dimensional trajectory points. By deploying it as a service, it supports high concurrency, making batch evaluation possible.

\phantomsection
\label{prompt:task decomposition}
\begin{tcolorbox}[colback=yellow!10!white, colframe=black!75!white, title=Prompt 5: Task Decomposition]
You are an assistant that can split complex pick and place tasks into simpler subtasks. Analyze the following task instruction and divide it into a list of subtasks.\\
Task instruction: \textit{\{instruction\}}.\\
Return the subtasks in JSON format as a list. If the task is simple and does not require splitting, return a list with a single subtask.\\
Each subtask should consist of a single pick and place operation.\\
It is normal for most tasks to have only one subtask.\\
Example output:
\renewcommand{\ttdefault}{cmtt}
\begin{verbatim}
{
    "subtasks":[
        "Pick the red apple and place it on the table.",
        "Pick the blue cup and place it next to the apple."
    ]
}
\end{verbatim}
\renewcommand{\ttdefault}{pcr}
\hrulefill\\
\textit{[Original Image]}
\end{tcolorbox}

\phantomsection
\label{prompt:SoM}
\begin{tcolorbox}[colback=yellow!10!white, colframe=black!75!white, title=Prompt 6: SoM]
You are an assistant designed for creating pick and place operations. The image shows a scene currently observed by the camera, and another image segmented the original image into different parts using a segmentation model with annotations on the corresponding masks. You need to understand the task instructions and select the object required for the current task.\\
Task instruction: \textit{\{instruction\}}.\\
Based on the above information, output the selected object in JSON format. If none of the masks include the object you need to select, output the number as -1 and the object name as "not\_found.\\
Example output:
\renewcommand{\ttdefault}{cmtt}
\begin{verbatim}
{
  "number": 3,
  "object_name": "apple",
  "color": "red"
}
\end{verbatim}
\renewcommand{\ttdefault}{pcr}
\hrulefill\\
\textit{[Original Image]}\textit{[Mask Annotated Image]}
\end{tcolorbox}

\phantomsection
\label{prompt:CtoF SoM}
\begin{tcolorbox}[colback=yellow!10!white, colframe=black!75!white, title=Prompt 7: Coarse-to-fine SoM]
You are an assistant designed for fine-grained part picking operations. The image shows the object that needs to be operated on, and another image segmented the original image into different parts using a segmentation model with annotations on the corresponding masks. You need to understand the task instructions and select the specific part of the object that needs to be picked.\\
Task instruction: \textit{\{instruction\}}.\\
Based on the above information, output the selected part in JSON format. If none of the masks include the part you need to select, output the number as -1 and the part name as "not\_found."\\
Example output:\\
\renewcommand{\ttdefault}{cmtt}
\begin{verbatim}
{
    "number": 3,
    "part_name": "handle",
    "color": "red"
}
\end{verbatim}
\renewcommand{\ttdefault}{pcr}
\hrulefill\\
\textit{[Croped Original Image]}\textit{[Croped Mask Annotated Image]}
\end{tcolorbox}

\phantomsection
\label{prompt:Grab Point Selection}
\begin{tcolorbox}[colback=yellow!10!white, colframe=black!75!white, title=Prompt 8: Grab Point Selection]
You need to grab the object: \textit{\{object\_name\}}. The image below are the original image and five sampled points.\\
Please select the most appropriate grab point by specifying the point number (1-5) and provide a brief reason.\\
Output the result in JSON format as follows:
\renewcommand{\ttdefault}{cmtt}
\begin{verbatim}
{
  "selected_point": 3,
  "reason": "Point 3 provides the most stable grip based on the object's orientation."
}
\end{verbatim}
\renewcommand{\ttdefault}{pcr}
\hrulefill\\
\textit{[Original Image]}\textit{[Point Annotated Image]}
\end{tcolorbox}

\phantomsection
\label{prompt:Grid-based Path Planning}
\begin{tcolorbox}[colback=yellow!10!white, colframe=black!75!white, title=Prompt 9: Grid-based Path Planning]
You are a motion planning agent. For the task I provide, you need to plan a series of waypoints and guide the robotic arm to the final position to complete the task.\\
For the task \textit{\{instruction\}}, with the previously identified object \textit{\{object\_name\}} and the grab points (marked in the image), you need to output the following series of details as described.\\
The whole path should begin with start point \textit{\{start\_point\}}\\
Each movement should include the grid cell number, the height above the table (0.1 to 0.5 meters), and the claw's orientation.\\
Output the list in JSON format where each item contains:\\
- grid\_number: e.g., `A1'\\
- height\_m: float value between 0.1 and 0.5\\
- claw\_orientation: one of [`up', `down', `left', `right', `front', `back']\\
Example Output:
\renewcommand{\ttdefault}{cmtt}
\begin{verbatim}
{
    "path":[\n'
        {"grid_number": "A1", "height_m": 0.3, "claw_orientation": "down"},
        {"grid_number": "B3", "height_m": 0.2, "claw_orientation": "front"},
        ...
    ]
}
\end{verbatim}
\renewcommand{\ttdefault}{pcr}
\hrulefill\\
\textit{[Original Image]}\textit{[Grid Annotated Image]}
\end{tcolorbox}

\phantomsection
\label{prompt:Point-to-Point Path Planning}
\begin{tcolorbox}[colback=yellow!10!white, colframe=black!75!white, title=Prompt 10: Point-to-Point Path Planning]
You are an assistant that selects a single point for the robot to move to.
For the task \textit{\{instruction\}}, with the previously identified object \textit{\{object\_name\}} and the grab points (marked in the image), you need to output the grid cell label to place the object at.\\
Based on the original image and the grid overlay, please select a grid cell label that represents the target point (where to place the object) for the robot.\\
Output the selected point in JSON format as follows:
\renewcommand{\ttdefault}{cmtt}
\begin{verbatim}
{
    "selected_point": {"type": "grid", "label": "C4"}
}
\end{verbatim}
\renewcommand{\ttdefault}{pcr}
\hrulefill\\
\textit{[Original Image]}\textit{[Grid Annotated Image]}
\end{tcolorbox}

\phantomsection
\label{prompt:Task completion checker}
\begin{tcolorbox}[colback=yellow!10!white, colframe=black!75!white, title=Prompt 11: Task completion checker]
You are an assistant that checks whether a task is completed. For the given instruction, the provided image describes the current scene, and you need to descibe the given scene and determine whether the task is finished.\\
Task instruction: \textit{\{instruction\}}.\\
Example output:
\renewcommand{\ttdefault}{cmtt}
\begin{verbatim}
{
    "scene_description": "The description of the scene",
    "finished": true or false
}
\end{verbatim}
\renewcommand{\ttdefault}{pcr}
\hrulefill\\
\textit{[Original Image]}
\end{tcolorbox}

\section{Implementation Details about Learning-based Models} \label{sec: Implementation Details about Learning-based Models}
Both GR-1 and ACT are trained on 1K trajectories per setting. For cross-scene training, each scene provides 1K episodes. The model predicts images from static and gripper cameras, forecasts the next 3 steps, and executes 1 step, using an input sequence length of 10. Training uses dropout, AdamW with cosine decay, a batch size of 512, and a learning rate of 2e-3. GR-1 is trained from scratch without pre-training.

\section{Failure Case Visualization of Modular Manipulation System} \label{sec: failure case visualization}

\begin{wrapfigure}{r}{0.35\textwidth}
  \centering
    \includegraphics[width=1.0\linewidth]{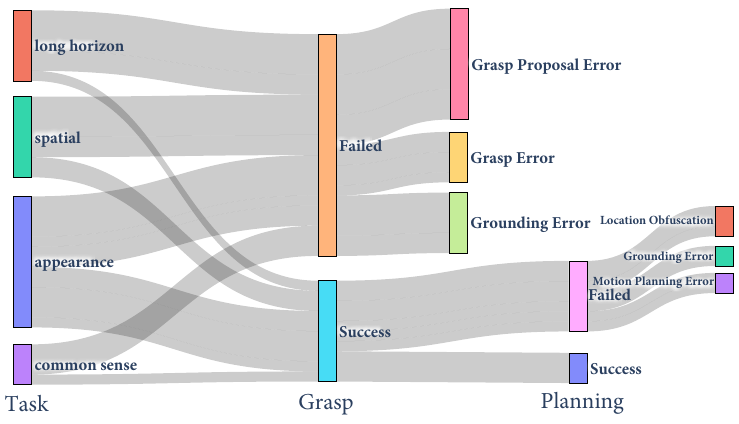}
    \caption{\textbf{Analysis of Failure Cases.} The figure categorizes failures by task type and identifies causes within grasp and planning.}
  \label{fig: failure case}
\end{wrapfigure}

We present a detailed analysis of agent failures, as illustrated in Figure A-~\ref{fig: failure case}. Tasks are categorized into four types: spatial, appearance, common sense, and long horizon. Grasp Failures are classified into three subcategories: (1) Grounding Error, resulting from incorrect masks generated by the Scene Object Model (SoM); (2) Grasp Proposal Error, caused by ungraspable poses identified by Context-to-Frame (CtoF) SoM or AnyGrasp; and (3) Grasp Error, arising from collisions or other operational issues. Similarly, Planning Failures are divided into: (1) Grounding Error, due to selecting an incorrect anchor object; (2) Location Obfuscation, involving confusion over placement specifications (\eg, distinguishing ``on'' from ``near''); and (3) Motion Planning Error, where the planned route cannot be executed.

\begin{figure}[h]
    \centering
    \includegraphics[width=1.0\linewidth]{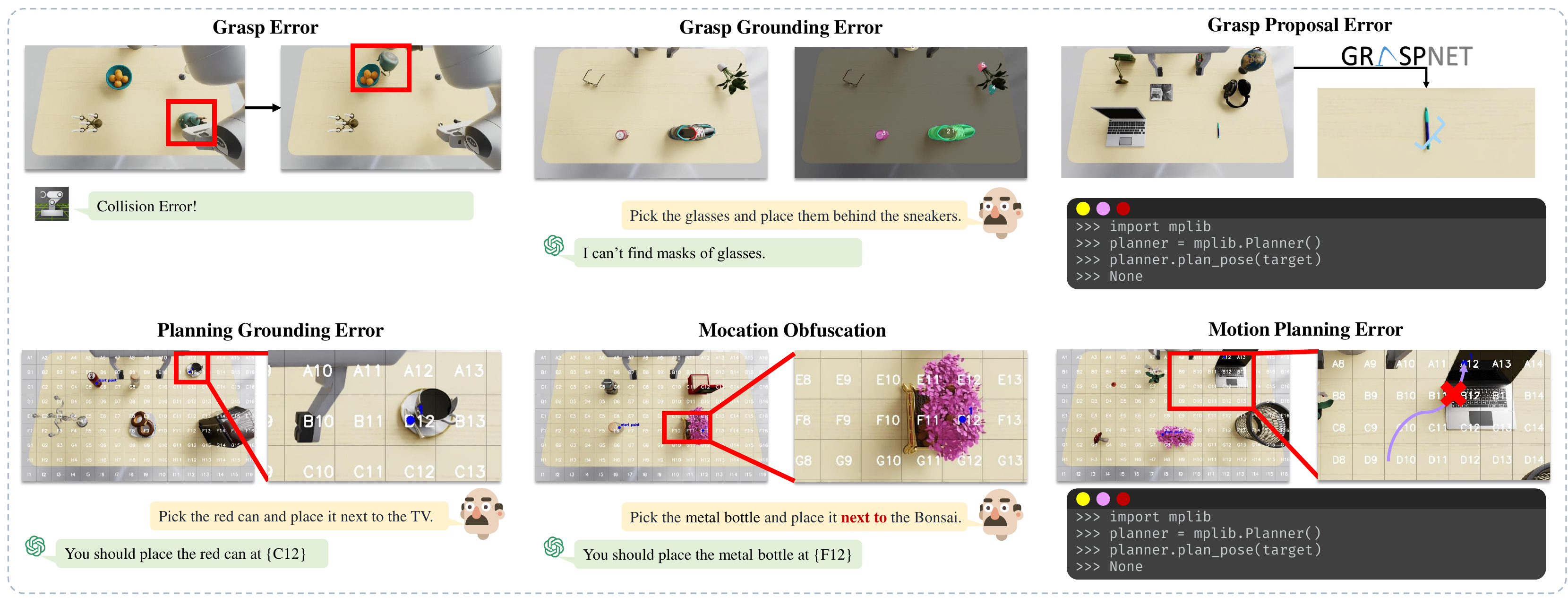}
    \caption{\textbf{Failure Cases Visualization.} Grasp Failures are classified into three subcategories: (1) Grounding Error, resulting from incorrect masks generated by the Scene Object Model (SoM); (2) Grasp Proposal Error, caused by ungraspable poses identified by Context-to-Frame (CtoF) SoM or AnyGrasp; and (3) Grasp Error, arising from collisions or other operational issues. Similarly, Planning Failures are divided into: (1) Grounding Error, due to selecting an incorrect anchor object; (2) Location Obfuscation, involving confusion over placement specifications (e.g., distinguishing ``on'' from ``near''); and (3) Motion Planning Error, where the planned route cannot be executed.}
    \label{fig: failure cases}
\end{figure}

The six representative failure cases are presented in Figure A-~\ref{fig: failure cases}. These examples highlight the limitations of the current prompt-based baseline, revealing areas for improvement such as selecting more optimal grasp poses and accurately grounding the destination. Addressing these challenges will require advancements in future methods, such as incorporating a stronger grasp proposal model, adopting a more robust chain of thought, or implementing closed-loop reasoning to enhance the accuracy and stability of the model's outputs.

\section{Characterizing Sim-to-Real Gap} \label{sec: characterizing sim-to-real gap}

\begin{figure}[h]
    \centering
    \includegraphics[width=1.0\linewidth]{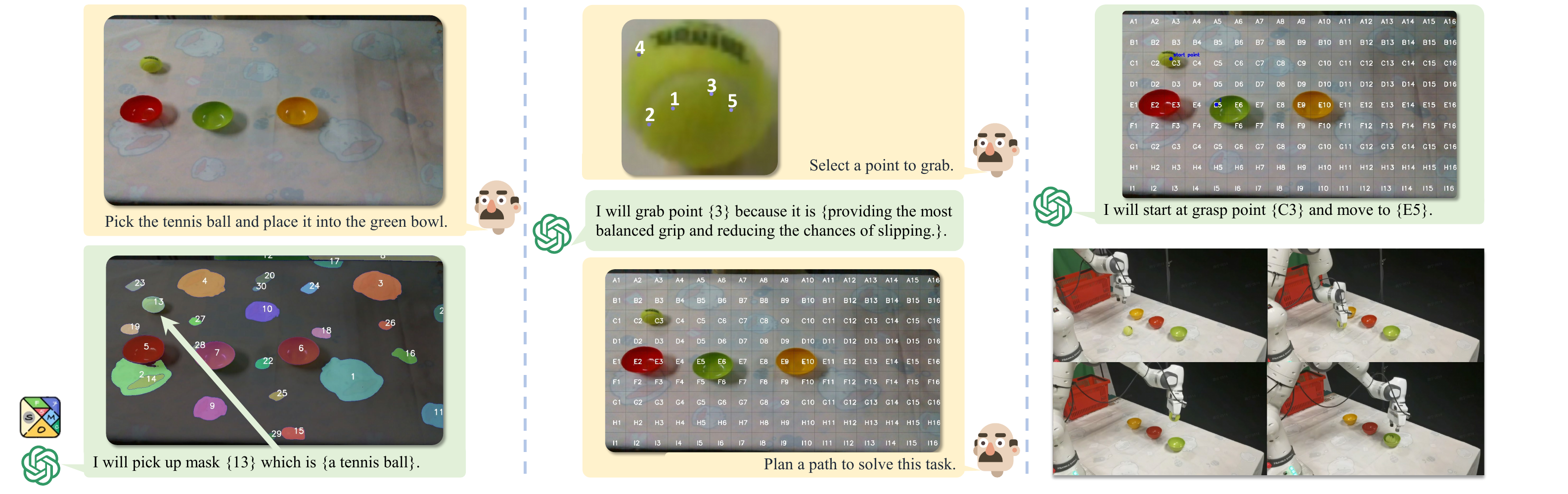}
    \caption{\textbf{Real-world deployment of modular manipulation system.}}

    \label{fig: prompt real}
\end{figure}

To demonstrate the realism of our environment, we deployed the prompt-based modular methods on a Franka robot in real-world scenarios. The chosen result, shown in Figure A-~\ref{fig: prompt real}, indicates that the baseline operates smoothly in the real environment with a measurable success rate. This suggests that our configured Isaac Sim environment closely mirrors the real world. Due to the costs associated with constructing real-world scenes, we leave further comparison between simulated and real environments as future work.

Despite the advancements achieved with \method in Isaac Sim, there still exists a notable sim-to-real gap between our simulation and the real-world environment. This gap manifests in the following two aspects:

\begin{itemize} 
    \item \textbf{Physical simulation}: Although Isaac Sim allows for the setting of physical parameters, including dynamic friction, static friction, restitution, density, etc., these settings demand extensive debugging and more detailed object annotations. Inappropriate friction coefficients may result in objects sticking or slipping during grasping, and unrealistic mass density can cause deviations in collision outcomes. 
    
    \item \textbf{Lighting simulation}: Isaac Sim provides an interface for setting material parameters, including roughness, metallic, ior, thin-walled, etc., and the objects themselves contain various materials. However, inconsistencies in object sources can lead to unrealistic settings, such as varying texture resolutions and reflectivity, potentially resulting in unrealistic simulation scenes. 
    
\end{itemize}

\end{document}